%% file: main.tex
\documentclass{article}

\usepackage{graphicx}
\usepackage{subcaption}
\usepackage{booktabs} 

\usepackage{tabularx}

\usepackage{url}
\usepackage{hyperref}

\usepackage{booktabs} 
\usepackage{multirow} 
\usepackage{wrapfig}    

\usepackage{xcolor}
\usepackage{transparent}
\usepackage{pifont}

\usepackage[toc,page]{appendix} 
\usepackage{titletoc}

\usepackage{longtable}
\usepackage{enumitem}

\usepackage{tabularx}
\usepackage{tikz}
\newcommand*\circled[1]{\tikz[baseline=(char.base)]{
  \node[shape=circle,draw,inner sep=1pt] (char) {#1};}}

\usepackage[table]{xcolor}

\usepackage[preprint]{icml2026}

\usepackage{amsmath}
\usepackage{amssymb}
\usepackage{mathtools}
\usepackage{amsthm}

\usepackage[capitalize,noabbrev]{cleveref}

\theoremstyle{plain}

\theoremstyle{definition}

\theoremstyle{remark}

\usepackage[textsize=tiny]{todonotes}

\icmltitlerunning{GenBench-MILP: A Benchmark Suite for MILP Instance Generation}

\usepackage{microtype}

\begin{document}

\twocolumn[
  \icmltitle{Are Your Generated Instances Truly Useful?
\texorpdfstring{\\}{ } GenBench-MILP: A Benchmark Suite for MILP Instance Generation}



  \icmlsetsymbol{equal}{*}

\begin{icmlauthorlist}
\icmlauthor{Yidong Luo}{sds}
\icmlauthor{Chenguang Wang}{sds}
\icmlauthor{Dong Li}{huawei}
\icmlauthor{Tianshu Yu}{sds}
\end{icmlauthorlist}

\icmlaffiliation{sds}{School of Data Science, The Chinese University of China, Shenzhen} 
\icmlaffiliation{huawei}{Huawei} 

\icmlcorrespondingauthor{Tianshu Yu}{yutianshu@cuhk.edu.cn}
  \icmlkeywords{Machine Learning, ICML}

  \vskip 0.3in
]



\begin{NoHyper}
\printAffiliationsAndNotice{}
\end{NoHyper}  

\begin{abstract}
The proliferation of machine learning-based methods for Mixed-Integer Linear Programming (MILP) instance generation has surged, driven by the need for diverse training datasets. However, a critical question remains: Are these generated instances truly useful and realistic? Current evaluation protocols often rely on superficial structural metrics or simple solvability checks, which frequently fail to capture the true computational complexity of real-world problems. To bridge this gap, we introduce GenBench-MILP, a comprehensive benchmark suite designed for the standardized and objective evaluation of MILP generators. Our framework assesses instance quality across four key dimensions: mathematical validity, structural similarity, computational hardness, and utility in downstream tasks. A distinctive innovation of GenBench-MILP is the analysis of solver-internal features—including root node gaps, heuristic success rates, and cut plane usage. By treating the solver's dynamic behavior as an expert assessment, we reveal nuanced computational discrepancies that static graph features miss. Our experiments on instance generative models demonstrate that instances with high structural similarity scores can still exhibit drastically divergent solver interactions and difficulty levels. By providing this multifaceted evaluation toolkit, GenBench-MILP aims to facilitate rigorous comparisons and guide the development of high-fidelity instance generators.  The code is available in \url{https://github.com/Aux-724/GenBench-MILP}
\end{abstract}

\input{introduction}   
\input{preliminaries}
\input{methodology}       
\input{experiments}   
\input{discussion}
\input{conclusion}    
\input{impact_statement}

\bibliography{references}
\bibliographystyle{icml2026}

\newpage
\appendix
\onecolumn
\clearpage
\input{appendix}

\end{document}

%% file: introduction.tex
\section{Introduction}

Mixed-Integer Linear Programming (MILP) is a fundamental optimization framework extending standard Linear Programming (LP) by incorporating integer variables, enabling the modeling of discrete decisions and logical conditions often intractable with purely continuous models \citep{achterberg2013, wolsey2020}. Its versatility in capturing complex relationships like fixed costs, mutual exclusivity, and indivisible entities makes MILP essential for practical decision-making in fields such as supply chain optimization, scheduling, financial modeling, and network design \citep{hugos2018, branke2015, mansini2015, alfalahy2017}.

Traditionally, MILP solver advancement relied on benchmark libraries like MIPLIB \citep{gleixner2021} and other real-world instance collections. \textit{However, these static benchmarks often lack the scalability, diversity, or controllable characteristics vital for modern research.} These limitations are particularly acute with the rise of machine learning (ML) in optimization, where learning-based approaches for tasks like branching strategy selection or algorithm configuration demand large, varied datasets that often exceed available collections \citep{bengio2021}. Consequently, a significant shift towards proactive \textbf{generation of MILP instances} has occurred \citep{bowly2020, geng2023, wang2023, guo2024, liu2024, yang2024, zeng2024, zhang2024}. This trend towards instance generation is motivated by multiple highlighted advantages: fulfilling the need for extensive datasets in ML applications; achieving greater instance diversity than found in existing libraries \citep{gleixner2021}; enabling control over computational difficulty for rigorous algorithm testing; facilitating the simulation of specific problem structures \citep{liu2024}; aiding solver testing and debugging \citep{gurobi2021}; and creating privacy-preserving surrogates for confidential real-world data.

\begin{figure*}[t]
    \centering
    \includegraphics[width=\textwidth]{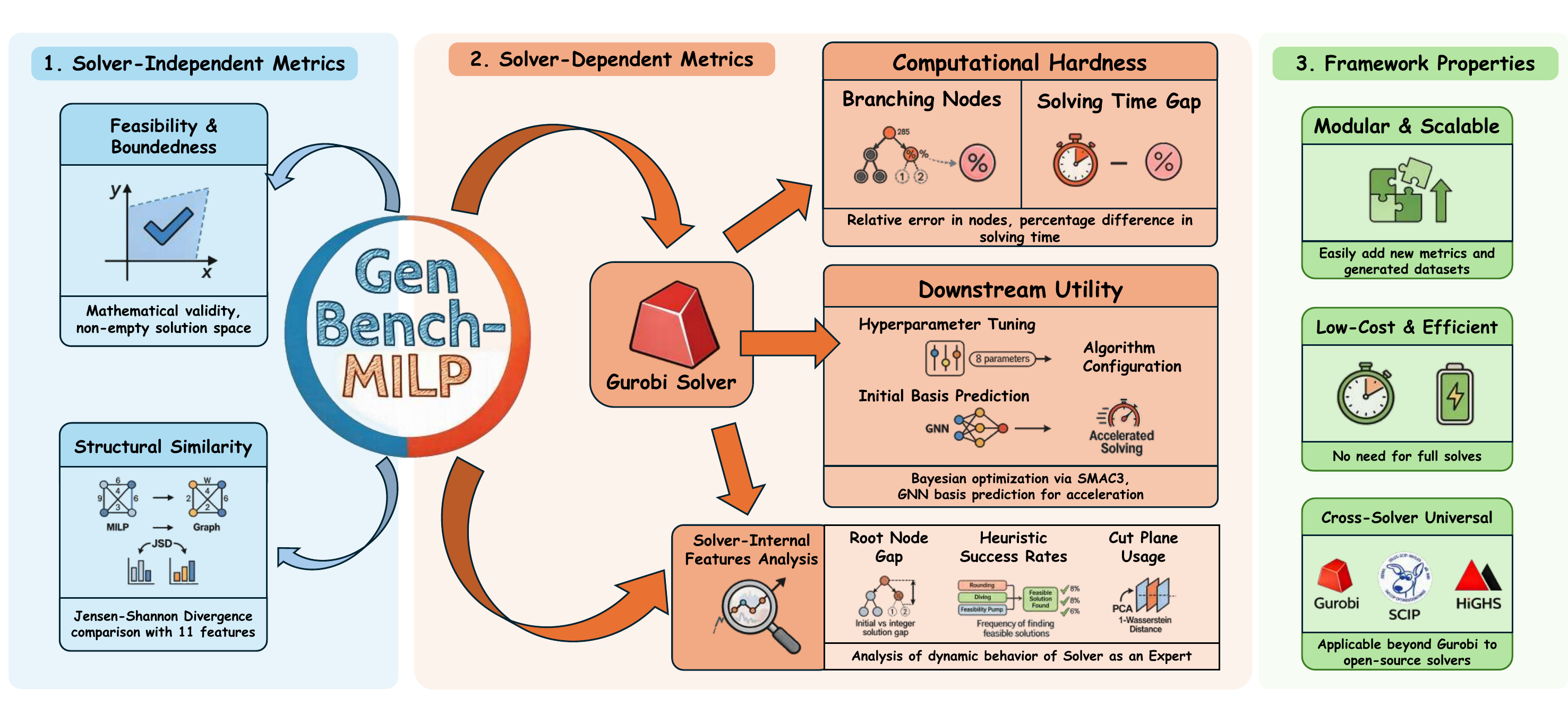} 
    \caption{\textbf{Architecture of GenBench-MILP}}
    \label{fig:eva_milp_architecture}
\end{figure*}

Methodologies for MILP instance generation have evolved in tandem. Early approaches featured parameterized generators for \textbf{specific problem classes} such as the traveling salesman problems~\citep{pilcher1992, vanderwiel1995}, set covering \citep{balas1980}, or quadratic assignment problems \citep{culberson2002}. Later techniques emphasized more \textbf{generalized feature-based descriptions} and \textbf{sampling} \citep{bowly2020, smith-miles2015}. Most recently, \textbf{deep learning based frameworks} like G2MILP \citep{geng2023}, MILP-StuDio \citep{liu2024}, and MILP-Evolve \citep{li2024} have gained prominence. These often represent MILP instances as graphs (e.g., bipartite graphs) and use \textbf{generative models} like variational autoencoders \citep{kipf2016} to learn and replicate specific structures from training data, \textit{marking a significant potential advancement but also introducing new evaluation challenges.} More information about related work could be found in Appendix~\ref{more_related_work}

Despite this progress in generation, a critical and multifaceted challenge remains: the fair, objective, and comprehensive evaluation of synthetic MILP instance ``quality''. Assessing generator outputs effectively requires considering several dimensions: \textbf{(I)} \textit{Ensuring fundamental solvability} (feasibility and boundedness) is crucial yet non-trivial \citep{wang2023, geng2023, liu2024}. \textbf{(II)}\textit{ Generated instances should ideally mirror real-world structural patterns} \citep{gleixner2021}, \textit{but mere resemblance may not capture true computational behavior.} \textbf{(III)} \textit{Instances must exhibit appropriate and controllable computational hardness}, typically evaluated via solver performance \citep{gurobi2021}, though achieving desired hardness levels consistently is a significant difficulty. Furthermore, \textit{current evaluation protocols} \citep{geng2023, bowly2020, zhang2024} \textit{often lack standardization, hindering robust comparisons and impeding progress.}

To address the lack of a standardized and comprehensive methodology for evaluating MILP instance generation techniques, this research introduces a novel benchmark framework. Our unified and extensible framework assesses instance quality across key dimensions -- mathematical validity, structural similarity, computational hardness, solver interaction patterns, and utility for downstream ML tasks. By establishing an objective standard, we aim to facilitate fairer comparisons, guide the development of higher-quality generators, and improve the reliability of research utilizing synthetic MILP data. The \textbf{practical viability} and \textbf{robustness }of our framework is comfirmed by our tests on the super hard dataset with a strict 120-second time limit, proving \textit{our solver-internal features are highly stable} and \textit{confirming the approach is both low-cost and efficient}. Our experiments on open-source solvers SCIP and HiGHS further confirm that while \textit{the framework is general, state-of-the-art solvers like Gurobi yield more precise internal features}. These findings not only validate our framework but also highlight a limitation in current DL-based generation approaches. While representing MILP instances as bipartite graphs has enabled progress in achieving structural similarity, this representation may struggle to fully capture the intricate constraint relationships and inherent mathematical properties crucial for MILP feasibility and hardness. We posit that future advancements might benefit from shifting focus from direct graph structure manipulation towards methods that more directly model the underlying mathematical structure or solution space characteristics. The proposed benchmark provides the necessary tools to rigorously evaluate such novel approaches.

Our specific contributions are four-fold: \textbf{(I)} We integrate both \textbf{Solver-Independent Metrics} (covering fundamental properties and structural resemblance) and \textbf{Solver-Dependent Metrics} (including computational characteristics and downstream task utility) for a holistic assessment. \textbf{(II)} We incorporate detailed analysis of \textbf{solver-internal features} (e.g., \textbf{root node gap}, \textbf{heuristic} success counts, \textbf{cut plane} profiles), using the solver's behavior as an expert assessment to reveal deeper computational similarities. \textbf{(III)} The framework is designed for \textbf{extendability}, featuring a modular structure to easily accommodate new metrics, datasets, and generation techniques over time. \textbf{(IV)} We provide the community with clear, operational metrics and quantitative criteria to objectively benchmark diverse generator types (e.g., rule-based, statistical, and machine learning-based), thereby steering future development.

%% file: preliminaries.tex
\section{Preliminaries} \label{sec:pre}

\textbf{MILP and its Bipartite Graph Representation}.
We consider the standard MILP problem:
\[
\min_{\mathbf{x} \in \mathbb{R}^n} \mathbf{\tilde{c}}^\top \mathbf{x}, \quad \textbf{s.t.} \quad \mathbf{A} \mathbf{x} \leq \mathbf{b}, \ x_i \in \mathbb{Z}, \ \forall i \in \mathcal{I}. 
\]

Here, \(\mathbf{x}\) is the vector of decision variables, \(\mathbf{\tilde{c}}\) contains the objective coefficients, \(\mathbf{A}\) is the constraint matrix, \(\mathbf{b}\) is the vector of constraint bounds, and \(\mathcal{I}\) identifies the integer variables.

Representing MILP instances as weighted bipartite graphs is an established practice in the relevant literature \citep{zhang2023survey, gasse_exact_2019, nair2021solving}. In this common representation, each MILP instance corresponds to a graph \(\mathcal{G}\) defined by three sets: constraint vertices \(\mathcal{C}\), variable vertices \(\mathcal{V}\), and edges \(\mathcal{E}\) connecting them. The constraint partition \(\mathcal{C} = \{c_1, \ldots, c_m\}\) includes one vertex \(c_i\) for each of the \(m\) constraints, where the vertex feature \(c_i\) typically represents the bias term, i.e., \(c_i = (b_i)\) (the \(i\)-th element of \(\mathbf{b}\) from the MILP formulation). The variable partition \(\mathcal{V} = \{v_1, \ldots, v_n\}\) contains one vertex \(v_j\) for each of the \(n\) variables, with the corresponding 9-dimensional feature vector \(\mathbf{v}_j\) encoding information such as the objective coefficient \(c_j\) (from the MILP vector \(\mathbf{c}\)), the variable type, and the bounds \(l_j, u_j\) (from the MILP vectors \(\mathbf{l}\) and \(\mathbf{u}\)). Edges in \(\mathcal{E}\) only exist between vertices from different partitions (\(c_i \in \mathcal{C}\) and \(v_j \in \mathcal{V}\)). An edge \(e_{i,j}\) connects \(c_i\) and \(v_j\) if the corresponding element \(A_{i,j}\) in the constraint matrix \(\mathbf{A}\) is non-zero; its associated feature \(e_{i,j}\) is described by this coefficient, i.e., \(e_{i,j} = (A_{i,j})\). If \(A_{i,j}\) is zero, the edge \(e_{i,j}\) is considered absent.

\textbf{Linear Programming Relaxation.} 
Given the MILP problem:
$$ z_{\text{MILP}} = \min_{\mathbf{x}} \{ \mathbf{\tilde{c}}^\top \mathbf{x} \mid \mathbf{A} \mathbf{x} \leq \mathbf{b}, x_i \in \mathbb{Z} \; \forall i \in \mathcal{I} \subseteq \{1, \dots, n\} \}. $$
The \textit{LP relaxation} is obtained by omitting the integrality constraints:
$$ z_{\text{LP}} = \min_{\mathbf{x}} \{ \mathbf{\tilde{c}}^\top \mathbf{x} \mid \mathbf{A} \mathbf{x} \leq \mathbf{b} \}. $$
Let $\mathcal{P} = \{\mathbf{x} \in \mathbb{R}^n \mid \mathbf{A} \mathbf{x} \leq \mathbf{b}\}$ be the feasible region of the LP relaxation. The optimal objective value of the LP relaxation provides a lower bound (for minimization problems) to the optimal objective value of the original MILP, i.e., $z_{\text{LP}} \leq z_{\text{MILP}}$.

\textbf{Duality Gap.} In the context of solving the MILP $\min \mathbf{\tilde{c}}^\top \mathbf{x}$, the \textit{duality gap} measures the difference between the objective value of the best known feasible integer solution, $z_{\text{incumbent}}$ (primal bound or upper bound, UB), and the best proven objective bound, $z_{\text{bound}}$ (dual bound or lower bound, LB), typically derived from LP relaxations ($z_{\text{bound}} \geq z_{\text{LP}}$). The absolute gap is $G_{\text{abs}} = z_{\text{incumbent}} - z_{\text{bound}}$ (assuming $z_{\text{incumbent}} \geq z_{\text{bound}}$). The relative gap, often used as a termination criterion, quantifies the remaining gap relative to one of the bounds, e.g.:
$$ G_{\text{rel}} = \frac{z_{\text{incumbent}} - z_{\text{bound}}}{\max(1, |z_{\text{incumbent}}|)}. $$
Optimality is proven when $z_\text{incumbent} = z_{\text{bound}}$, or $G_{\text{abs}}$ (or $G_{\text{rel}}$) is within a predefined tolerance $\epsilon \geq 0$.

\textbf{Feasibility} \quad Consider the MILP problem defined by constraints $\mathbf{A} \mathbf{x} \leq \mathbf{b}$ and $x_i \in \mathbb{Z}, \forall i \in \mathcal{I}$. The feasible region is the set $\mathcal{F} = \{\mathbf{x} \in \mathbb{R}^n \mid \mathbf{A} \mathbf{x} \leq \mathbf{b}, x_i \in \mathbb{Z} \; \forall i \in \mathcal{I}\}$. The MILP problem is \textit{feasible} if its feasible region $\mathcal{F}$ is non-empty, i.e., $\mathcal{F} \neq \emptyset$. A specific point $\hat{\mathbf{x}} \in \mathbb{R}^n$ is considered \textit{feasible} if it satisfies all constraints, i.e., $\hat{\mathbf{x}} \in \mathcal{F}$.

\textbf{Boundedness} \quad Given the MILP objective $\min_{\mathbf{x} \in \mathcal{F}} \mathbf{\tilde{c}}^\top \mathbf{x}$, the problem is \textit{bounded} if the optimal objective value $z^*$ is finite. This means $z^* = \inf_{\mathbf{x} \in \mathcal{F}} \mathbf{\tilde{c}}^\top \mathbf{x} > -\infty$. (For a maximization problem $\max_{\mathbf{x} \in \mathcal{F}} \mathbf{\tilde{c}}^\top \mathbf{x}$, boundedness would mean $z^* = \sup_{\mathbf{x} \in \mathcal{F}} \mathbf{\tilde{c}}^\top \mathbf{x} < +\infty$). If a problem is both feasible ($\mathcal{F} \neq \emptyset$) and bounded, a finite optimal objective value $z_{MILP}$ exists.

Additional preliminaries can be found in Appendix \ref{more preli}.

%% file: methodology.tex
\section{Methodology} \label{methods}

This section outlines the methodology employed to evaluate the quality and characteristics of generated MILP instances. We compare instances generated using reproductions of three open-source methods – G2MILP \citep{geng2023}, ACM-MILP \citep{guo2024}, and DIG-MILP \citep{wang2023} – against a baseline set of `original' instances. This baseline set, comprising Set Cover, Combinatorial Auction (CA), Capacitated Facility Location (CFL), and Independent Set (IS) problems, was synthesized using the Ecole library \citep{prouvost2020ecole} (the specific parameter settings used for this synthesis are detailed in the Appendix~\ref{sec:datasets}), while IP and LB come from two challenging real-world problem families used in ML4CO 2021 competition \citep{gasse2022ml4co}. To capture different facets of similarity and fidelity between the generated and original sets, we utilize a suite of metrics. These metrics are organized into two primary categories based on their reliance on the underlying optimization solver: \textbf{(I)} \textbf{Solver-Independent Metrics}, which assess inherent properties of the MILP instances themselves (e.g., feasibility, structure) without regard to any specific solution process; and \textbf{(II)} \textbf{Solver-Dependent Metrics}, which evaluate aspects intrinsically linked to the behavior and performance of a particular solver (primarily Gurobi) when applied to the instances. This dual approach allows for a robust assessment, considering both fundamental instance properties and their practical implications for optimization algorithms (See the whole framework in Figure~\ref{fig:eva_milp_architecture}).

\subsection{Datasets}
This study evaluates MILP instances from two sources: original instances (baselines and validation) and those from three generative models (G2MILP, ACM-MILP, DIG-MILP). Baselines include four synthetic datasets (SC, CA, CFL, IS) from Ecole \citep{prouvost2020ecole} (details in Appendix~\ref{sec:datasets}) and public benchmarks from the ML4CO Competition 2024 \citep{gasse2022ml4co} for pipeline validation. 
Due to the inherent specialization of current generative models to particular problem types, our comparative analysis specifically examines SC instances from G2MILP, CA instances from ACM-MILP and DIG-MILP, and IS instances from ACM-MILP and G2MILP. We emphasize that these specific combinations serve primarily as representative examples to demonstrate the utility of our metrics; the GenBench-MILP framework itself is dataset-agnostic, and its broader applicability to other distinct datasets and models has been confirmed (see Section~\ref{ml4co}).

\subsection{Solver-Independent Metrics}

\subsubsection{Feasibility Ratio}
This metric measures the proportion of instances that are both \textbf{feasible} and \textbf{bounded}. Most methods maintain near-perfect feasibility across datasets, with only minor exceptions (e.g., G2MILP on MIS at $\eta=0.10$). Detailed results are given in Table~\ref{tab:feasibility_all} in the Appendix.

\subsubsection{Structural Similarity}

\begin{table}[ht]
    \centering
    \small 
    \setlength{\tabcolsep}{5pt} 
    \caption{\textbf{Structural Similarity.} The similarity score is derived from Jensen-Shannon Divergence (higher is better). The \textbf{Masked Ratio ($\eta$)} denotes the proportion of the original instance's structure (e.g., constraint nodes in G2MILP or constraints in ACM-MILP) that is masked, corrupted, or selected for reconstruction during the generation process, serving as a control parameter for the modification magnitude.}
    \label{tab:similarity}
    \begin{tabular}{@{}llcccc@{}}
        \toprule
        Problem & Model    & \multicolumn{4}{c}{Similarity at Masked Ratio ($\eta$)} \\
        \cmidrule(lr){3-6}
                &          & 0.01   & 0.05    & 0.10    & 0.20    \\
        \midrule
        CA      & ACM-MILP & ---    & 0.889   & 0.867   & 0.892   \\
        CA      & DIG-MILP & 0.861  & 0.860   & 0.879   & ---     \\ 
        \midrule
        IS      & ACM-MILP & ---    & 0.734   & 0.731   & 0.699   \\
        IS      & G2MILP   & \cellcolor{gray!20}0.486  & \cellcolor{gray!20}0.473   & \cellcolor{gray!20}0.466   & ---     \\ 
        \midrule
        SC      & G2MILP   & \cellcolor{orange!30}\textbf{0.990 } & \cellcolor{orange!30}\textbf{0.950}   & \cellcolor{orange!30}\textbf{0.921 }  & ---     \\
        \bottomrule
    \end{tabular}
\end{table}

To quantitatively assess the \textbf{structural similarity} between two sets of MILP instances, following \citep{geng2023}, we implemented a feature-based comparison method. Each instance is first converted into a graph representation, from which a vector of \textbf{11 predefined structural features} is extracted. For each feature, we compute the \textbf{Jensen--Shannon Divergence (JSD)} between the distributions of the two sets and transform it into a similarity score $S_i = 1 - \frac{JSD_i}{\log(2)}$, where a score of 1 indicates identical distributions and 0 indicates maximal divergence. The overall structural similarity between the two sets is the arithmetic mean of these feature scores. Overall, the JSD-based metric (Table~\ref{tab:similarity}) indicates that ACM-MILP and DIG-MILP closely match the structural characteristics of CA data (around 0.86–0.89), while ACM-MILP achieves moderate similarity on IS (around 0.70) and G2MILP is lower (around 0.47). For Set Cover, G2MILP performs best, reaching very high similarity (around 0.92–0.99), especially at lower mask ratios.

\subsection{Solver-Dependent Metrics}

\subsubsection{Branching nodes}
Computational hardness is measured by the number of \textbf{branch-and-bound nodes} explored by Gurobi, using the relative error\(\left| \frac{\sum Nodes_{\text{gen}} - \sum Nodes_{\text{train}}}{\sum Nodes_{\text{train}}} \right|\!\times\!100\%.\) As shown in Table~\ref{tab:branching_nodes} (Appendix F), hardness varies strongly by model and problem type: G2MILP on IS can exceed 50,000 \% RE and often times out, ACM-MILP on CA reaches over 26,000 \% RE, while DIG-MILP and other settings remain closer to baseline.

\begin{table}[htbp]
  \centering
  \caption{\textbf{Branching Node Statistics and Relative Error across Problem Types.} Statistics are calculated over 1000 instances per source/ratio. \textbf{Original} refers to the original training sets (Total Nodes: IS=14,255; CA=18,488; SC=46,408). 'Time Limit Hits' indicates premature termination. Relative Error (RE) compares generated total nodes to baseline total nodes. $\eta$ denotes the mask ratio.}
  \label{tab:branching_nodes} 
  \resizebox{\linewidth}{!}{%
  \begin{tabular}{@{}l l c r r r r r r@{}} 
    \toprule
    Problem & Source & \shortstack{Mask\\Ratio\\($\eta$)} & \shortstack{Mean\\Nodes} & \shortstack{Median\\Nodes} & \shortstack{Std\\Dev} & \shortstack{Max\\Nodes} & \shortstack{Time Limit\\Hits} & \shortstack{Relative\\Error (\%)} \\
    \midrule
    IS & Original & N/A   & 14.3   & 1.0    & 49.1    & 1,056  & 0   & {---} \\
    \cmidrule(lr){2-9} 
        & G2MILP & 0.01 & 26.2   & 1.0    & 60.1    & 724    & 0   & 84.1 \\
        & G2MILP & 0.05 & 765.7  & 175.0  & 1,941.6 & 16,314 & 5   & 5,271.6 \\
        & G2MILP & 0.10 & 8,386.4 & 8,330.0 & 1,219.5 & 11,789 & 982 & \cellcolor{gray!20}58,730.0 \\
    \cmidrule(lr){2-9} 
        & ACM-MILP & 0.05 & 8.8    & 1.0    & 36.8    & 572    & 0   & 38.3 \\
        & ACM-MILP & 0.10 & 10.9   & 1.0    & 63.8    & 1,344  & 0   & \cellcolor{orange!30}\textbf{23.4} \\
        & ACM-MILP & 0.20 & 12.5   & 1.0    & 78.0    & 1,977  & 0   & \cellcolor{orange!30}\textbf{12.3} \\
    \midrule
    CA  & Original & N/A   & 18.5    & 1.0     & 48.1     & 514    & 0   & {---} \\
    \cmidrule(lr){2-9} 
        & ACM-MILP & 0.05 & 6,556.0 & 1,661.5 & 9,926.3  & 51,913 & 138 & \cellcolor{gray!20}35,360.0 \\
        & ACM-MILP & 0.10 & 6,504.9 & 1,547.0 & 9,962.8  & 54,385 & 120 & \cellcolor{gray!20}35,084.3 \\
        & ACM-MILP & 0.20 & 4,873.6 & 962.5   & 8,021.7  & 43,036 & 75  & 26,259.8 \\
    \cmidrule(lr){2-9} 
        & DIG-MILP & 0.01 & 34.1 & 1.0 & 76.2 & 799 & 0 & 84.4 \\
        & DIG-MILP & 0.05 & 34.3 & 1.0 & 76.4 & 812 & 0 & 85.6 \\
        & DIG-MILP & 0.10 & 37.8 & 1.0 & 83.5 & 902 & 0 & 104.3 \\
    \midrule
    SC  & Original & N/A   & 46.4   & 1.0    & 298.6   & 8,135  & 0   & {---} \\
    \cmidrule(lr){2-9} 
        & G2MILP & 0.01 & 85.4   & 1.0    & 536.5   & 14,592 & 0   & 84.0 \\
        & G2MILP & 0.05 & 79.6   & 1.0    & 430.9   & 10,643 & 0   & 71.6 \\
        & G2MILP & 0.10 & 63.7   & 1.0    & 269.8   & 4,713  & 0   & \cellcolor{orange!30}\textbf{37.3} \\
    \bottomrule
  \end{tabular}%
  } 
\end{table}

\subsubsection{Solving Time Gap}
This metric measures the percentage difference in average solving times between generated and original instances, where smaller gaps indicate closer solver-conditional hardness. Table~\ref{tab:solving_time_gap} in Appendix shows that G2MILP for SC keeps gaps modest (\(\approx\)11–22\%), DIG-MILP for CA remains near baseline (\(\approx\) 20–29\%), while ACM-MILP can greatly alter hardness (e.g., 2400\%+ increase for CA or \(\sim\)47–49\% faster for IS).

\subsubsection{Hyperparameter Tuning}

Automatically optimizing solver hyperparameters is known to be crucial for achieving peak performance on complex algorithms like MILP solvers \citep{hutter2011sequential}. Therefore, we employed the Sequential Model-based Algorithm Configuration (\textbf{SMAC3}) framework \citep{JMLR:v23:21-0888}, which utilizes Bayesian optimization. Following the approach in \citep{liu2024}, the tuning process targeted \textbf{8 key Gurobi parameters}. These parameters were selected as they govern diverse and fundamental components of the MILP solution process, including primal heuristics, search strategy focus, branching decisions, presolving routines, cutting plane generation, and node LP solution methods. The significant impact of these core solver components, and thus the parameters controlling them, on overall performance is well-documented in the MILP literature \citep[e.g.,][]{lodi2017algorithmic, achterberg2009scip, berthold2012primal}. The objective function for SMAC3 was the minimization of the mean wall-clock solve time across the instances within a designated tuning set. The primary goal of this hyperparameter tuning experiment was to evaluate the \textbf{generalization} capability of the optimized Gurobi configurations using generated MILP instances. SMAC3 tuning (Table~\ref{tab:solving_time_improvement}) gives model- and dataset-specific gains. ACM-MILP (IS) decreases about \textbf{10\%} at $\eta=0.1$. DIG-MILP (CA) consistently cuts time by about \textbf{61\%}, and G2MILP (SC) by about \textbf{44\%}, while the original SC set improves only about \textbf{3\%}.

\vspace{-1ex}

\begin{table}[htbp]
    \centering
    \caption{\textbf{Solving time improvement after Gurobi hyperparameter tuning}. Compares the average wall-clock time (seconds) on the test set for the default configuration versus the best configuration found by SMAC3.}
    \label{tab:solving_time_improvement}
    
    \resizebox{\linewidth}{!}{%
    \begin{tabular}{@{}llc ccc@{}}
    \toprule
    Dataset & Source     & $\eta$ & Default(s) & Best(s) & Improv. (\%) \\
    \midrule
    \multirow{4}{*}{IS} 
            & ACM-MILP         & 0.05   & 0.339            & 0.370         & \cellcolor{gray!20}-9.17            \\
            &                  & 0.1    & 0.351            & 0.280         & 20.25            \\
            &                  & 0.2    & 0.339            & 0.377         & \cellcolor{gray!20}-11.13           \\
            & Original         & ---    & 0.336            & 0.366         & 8.98             \\
    \midrule
    \multirow{4}{*}{CA} 
            & DIG-MILP         & 0.01   & 0.107            & 0.041         & \cellcolor{orange!30}\textbf{61.35}         \\
            &                  & 0.05   & 0.109            & 0.042         & \cellcolor{orange!30}\textbf{61.63}         \\
            &                  & 0.1    & 0.108            & 0.042         & \cellcolor{orange!30}\textbf{61.38}          \\
            & Original         & ---    & 0.107            & 0.042         & \cellcolor{orange!30}\textbf{61.26}          \\
    \midrule
    \multirow{3}{*}{SC} 
            & G2MILP           & 0.01   & 0.211            & 0.118         & 43.97            \\
            &                  & 0.05   & 0.211            & 0.118         & 44.06            \\
            & Original         & ---    & 0.229            & 0.221         & 3.31             \\
    \bottomrule
    \end{tabular}%
    }
\end{table}

\subsubsection{Initial Basis Prediction}
Following the methodology of \citep{fan2023}, we investigated predicting an \textbf{initial basis} for MILP relaxations using a GNN. While adhering to the principles outlined by Fan et al., our implementation was developed independently from scratch. For this purpose, a GNN model was trained specifically on generated MILP instances to predict the basis status (basic or non-basic at lower/upper bounds) for variables and slack variables associated with the MILP relaxation. The GNN's predictions were subsequently refined into a numerically stable basis using established techniques. To evaluate the effectiveness of training on generated data versus original data for this task, we measured the impact of using the initial basis predicted by the GNN on Gurobi's performance metrics (solving time, runtime). This was compared against Gurobi's default setting and potentially a model trained on original data, using unseen test instances. The validity and effectiveness of our re-implemented approach are substantiated by the experimental outcomes. Resuls are in Table~\ref{tab:basis_results}.

\begin{table}[htbp]
 \caption{\textbf{Impact of Generative Data Augmentation on GNN Initial Basis Prediction.} This table evaluates the solving time improvement achieved by initializing Gurobi with a GNN-predicted basis compared to the default initialization. The GNN models were trained either on the \textbf{Original} dataset alone or on the original dataset augmented with synthetic instances generated by different \textbf{Sources}. All performance metrics (average runtime in seconds) are measured on the unseen original test set. \textbf{Default}: Gurobi's default runtime; \textbf{Best}: Runtime using the GNN-predicted basis; \textbf{Improv.}: Percentage reduction in runtime relative to the default.}
 \label{tab:basis_results} 
 \centering
 \resizebox{\linewidth}{!}{%
  \begin{tabular}{@{}lllrrr@{}}
  \toprule
  Dataset & Source & $\eta$ & Default (s) & Best (s) & Improv. (\%) \\
  \midrule
  CA & ACM-MILP & 0.05 & 0.078 & 0.077 & 2.6\% \\
     &          & 0.10 & 0.079 & 0.076 & 2.6\% \\
     &          & 0.20 & 0.078 & 0.075 & 4.0\% \\
  \cmidrule(lr){2-6} 
     & DIG-MILP & 0.05 & 0.079 & 0.076 & 3.5\% \\
     &          & 0.10 & 0.080 & 0.076 & 4.1\% \\
  \cmidrule(lr){2-6} 
     & Original & --- & 0.078 & 0.077 & 2.5\% \\
  \midrule
  IS & ACM-MILP & 0.10 & 0.236 & 0.239 & \cellcolor{gray!20}-0.2\% \\
  \cmidrule(lr){2-6} 
      & G2MILP  & 0.05 & 0.237 & 0.240 & \cellcolor{gray!20}-0.6\% \\
  \cmidrule(lr){2-6} 
      & Original & --- & 0.236 & 0.237 & \cellcolor{gray!20}0.2\% \\
  \midrule
  SC & G2MILP & 0.01 & 0.236 & 0.221 & \cellcolor{orange!30}\textbf{10.6\%} \\
     &        & 0.05 & 0.240 & 0.229 & \cellcolor{orange!30}\textbf{9.1\%} \\
     &        & 0.10 & 0.242 & 0.224 & \cellcolor{orange!30}\textbf{12.4\%} \\
  \cmidrule(lr){2-6} 
     & Original & --- & 0.236 & 0.226 & \cellcolor{orange!30}\textbf{9.1\%} \\
  \bottomrule
  \end{tabular}%
 }
\end{table}

\subsubsection{Solver-Internal Features}
We introduce a novel, computationally efficient methodology to evaluate MILP instance fidelity using \textbf{solver-internal features}: metrics from Gurobi's deterministic solving process. We first validated the solver-internal–feature methodology via a split-half experiment (see Appendix~\ref{validation}). Figure~\ref{fig:is_sif} shows one representative result, where random original-set subsets exhibit very low \textbf{1-Wasserstein distances} (e.g. 0.130 for heuristic, 0.188 for root-node gap, 0.232 for cut-plane profiles), confirming its ability to capture authentic inter-set similarity. To further demonstrate practical viability, we applied the method to the \textbf{hard} ML4CO \textit{item\_placement} dataset under a strict 120-second limit (see more in Section~\ref{ml4co}. This shows that we do not need to fully solve the instance to extract the features. Solver-internal features extracted from Gurobi proved stable and inexpensive to compute. What's more, comparison with experiments of open-source solvers \textbf{SCIP} and \textbf{HiGHS} (see more in Section~\ref{crosssolver}) indicates that, although the framework is general, state-of-the-art solvers like Gurobi provide more precise internal metrics. 

\begin{table}[htbp]
\centering
\caption{\textbf{Comparison of Root Node Gap Distributions}}
\label{tab:rootgap}
\resizebox{\linewidth}{!}{%
    \setlength{\tabcolsep}{3pt} 
    \begin{tabular}{@{}ll c cc cc c@{}} 
    \toprule
     & & & \multicolumn{2}{c}{Mean Gap (\%)} & \multicolumn{2}{c}{Std Dev (\%)} & \\
    \cmidrule(lr){4-5} \cmidrule(lr){6-7}
    Problem & Model    & $\eta$ & Gen & Original & Gen & Original & $W_1$ Dist. \\
    \midrule
    \multirow{6}{*}{CA} 
            & \multirow{3}{*}{DIG-MILP} & 0.01 & 1.11 & 2.47 & 1.22 & 2.60 & 1.3001 \\
            &           & 0.05 & 1.09 & 2.47 & 1.17 & 2.60 & 1.3787 \\
            &           & 0.10 & 1.18 & 2.47 & 1.26 & 2.60 & 1.2982 \\
    \cmidrule(l){2-8} 
            & \multirow{3}{*}{ACM-MILP} & 0.05 & 7.93 & 2.47 & 3.85 & 2.60 & \cellcolor{gray!20}\textbf{5.4638} \\
            &           & 0.10 & 7.49 & 2.47 & 3.60 & 2.60 & 5.0187 \\
            &           & 0.20 & 9.54 & 2.47 & 4.40 & 2.60 & \cellcolor{gray!20}\textbf{7.0666} \\
    \midrule
    \multirow{3}{*}{IS} 
            & \multirow{3}{*}{ACM-MILP} & 0.05 & 2.65 & 3.42 & 1.82 & 1.34 & 0.8818 \\
            &           & 0.10 & 2.66 & 3.42 & 1.81 & 1.34 & 0.8646 \\
            &           & 0.20 & 2.45 & 3.42 & 1.80 & 1.34 & 1.0388 \\
    \midrule
    \multirow{3}{*}{SC} 
            & \multirow{3}{*}{G2MILP} & 0.01 & 8.25 & 7.61 & 4.19 & 4.12 & 0.6690 \\
            &           & 0.05 & 7.75 & 7.61 & 4.27 & 4.12 & \cellcolor{orange!30}\textbf{0.2294} \\
            &           & 0.10 & 8.05 & 7.61 & 4.12 & 4.12 & \cellcolor{orange!30}\textbf{0.5103} \\
    \bottomrule
    \end{tabular}%
}
\end{table}

With fixed Gurobi settings, we extracted:\textbf{ Root Node Gap}, \textbf{Heuristic Success Count}, and \textbf{Cut Plane Usage vectors}. Distributions for Root Node Gap and Heuristic Success Count were compared using the 1-Wasserstein distance (\(W_1\)). For Cut Plane Usage, per-instance vectors were preprocessed, underwent PCA, and distributions of scores on dominant principal components were then compared using the \(W_1\) distance.

\begin{table}[htbp]
\centering
\caption{\textbf{Comparison of Heuristic Success Frequency Distributions}}
\label{tab:heuristic_comparison}
\resizebox{\linewidth}{!}{%
    \setlength{\tabcolsep}{3pt} 
    \begin{tabular}{@{}ll c cc cc c@{}}
    \toprule
    & & & \multicolumn{2}{c}{Gen. Stats} & \multicolumn{2}{c}{Base Stats} & \\
    \cmidrule(lr){4-5} \cmidrule(lr){6-7}
    Problem & Model & $\eta$ & Mean & Std & Mean & Std & $W_1$ Dist. \\
    \midrule
    \multirow{6}{*}{CA} 
            & \multirow{3}{*}{ACM-MILP} & 0.05 & 2.549 & 0.500 & 2.115 & 0.379 & 0.434 \\
            &          & 0.10 & 2.552 & 0.501 & 2.115 & 0.379 & 0.437 \\
            &          & 0.20 & 2.457 & 0.500 & 2.115 & 0.379 & 0.342 \\
    \cmidrule(l){2-8} 
            & \multirow{3}{*}{DIG-MILP} & 0.01 & 2.238 & 0.451 & 2.115 & 0.379 & 0.135 \\
            &          & 0.05 & 2.241 & 0.439 & 2.115 & 0.379 & 0.126 \\
            &          & 0.10 & 2.258 & 0.440 & 2.115 & 0.379 & 0.143 \\
    \midrule
    \multirow{3}{*}{IS} 
            & \multirow{3}{*}{ACM-MILP} & 0.05 & 1.662 & 0.473 & 18.883 & 12.112 & \cellcolor{gray!20}17.221 \\
            &          & 0.10 & 1.746 & 0.435 & 18.883 & 12.112 & \cellcolor{gray!20}17.137 \\
            &          & 0.20 & 1.849 & 0.358 & 18.883 & 12.112 & \cellcolor{gray!20}17.034 \\
    \midrule
    \multirow{3}{*}{SC} 
            & \multirow{3}{*}{G2MILP}   & 0.01 & 2.619 & 0.533 & 2.702 & 0.491 & \cellcolor{orange!30}\textbf{0.083} \\
            &          & 0.05 & 2.600 & 0.548 & 2.702 & 0.491 & \cellcolor{orange!30}\textbf{0.102} \\
            &          & 0.10 & 2.598 & 0.541 & 2.702 & 0.491 & \cellcolor{orange!30}\textbf{0.104} \\
    \bottomrule
    \end{tabular}%
}
\end{table}

\textbf{Results}\quad Table~\ref{tab:rootgap} presents \(W_1\) distances for root gap distributions (lower \(W_1\) indicates higher similarity). G2MILP highly replicated SC root gaps (e.g., \(W_1 = 0.2294\) at \(\eta = 0.05\)). In contrast, DIG-MILP and ACM-MILP significantly diverged for CA. ACM-MILP showed moderate success for IS at lower \(\eta\), diminishing as \(\eta\) increased.

\begin{table}[ht]
    \centering
    \small
    \setlength{\tabcolsep}{5pt}
    \caption{\textbf{Comparison of Cutting Plane Usage Distributions}}
    \label{tab:cutplane_condensed}
    \begin{tabular}{@{}ll c ccc@{}}
    \toprule
     & & & \multicolumn{3}{c}{Wasserstein Distance} \\
    \cmidrule(lr){4-6}
    Problem & Model & $\eta$ & PC1 & PC2 & PC3 \\
    \midrule
    \multirow{6}{*}{CA}
            & \multirow{3}{*}{ACM-MILP} & 0.05 & \cellcolor{gray!20}1.5570 & 0.8352 & 0.5349 \\
            &          & 0.10 & \cellcolor{gray!20}1.4837 & 0.7155 & 0.3570 \\
            &          & 0.20 & \cellcolor{gray!20}1.6889 & 0.3166 & 0.3731 \\
    \cmidrule(l){2-6}
            & \multirow{3}{*}{DIG-MILP} & 0.01 & \cellcolor{gray!20}1.4711 & 0.6673 & 0.3601 \\
            &          & 0.05 & \cellcolor{gray!20}1.4868 & 0.6826 & 0.3047 \\
            &          & 0.10 & \cellcolor{gray!20}1.5392 & 0.4394 & 0.5988 \\
    \midrule
    \multirow{3}{*}{IS}
            & \multirow{3}{*}{ACM-MILP} & 0.05 & 0.8944 & 0.4986 & 0.2517 \\
            &          & 0.10 & 0.9354 & 0.3942 & 0.0963 \\
            &          & 0.20 & 1.0097 & 0.5735 & 0.1199 \\
    \midrule
    \multirow{3}{*}{SC}
            & \multirow{3}{*}{G2MILP}   & 0.01 & \cellcolor{orange!30}\textbf{0.0707} & 0.1271 & 0.1124 \\
            &          & 0.05 & \cellcolor{orange!30}\textbf{0.1714} & 0.1425 & 0.0677 \\
            &          & 0.10 & \cellcolor{orange!30}\textbf{0.1374} & 0.1028 & 0.1375 \\
    \bottomrule
    \end{tabular}
\end{table}

For primal heuristic success distributions (Table~\ref{tab:heuristic_comparison}, \(W_1\) distance), DIG-MILP (e.g., \(W_1 \approx 0.14\)) and ACM-MILP (e.g., \(W_1 \approx 0.34 - 0.44\)) achieved good similarity for CA problems. G2MILP (SC) demonstrated remarkable similarity (\(W_1 \approx 0.08 - 0.10\)). Conversely, ACM-MILP (IS) instances diverged significantly (\(W_1 \approx 17.0 - 17.2\)).

For cutting plane usage (Table~\ref{tab:cutplane_condensed}), G2MILP (SC) achieved the highest similarity (e.g., PC1 \(W_1 \approx 0.07 - 0.17\)). In contrast, ACM-MILP and DIG-MILP for CA instances showed significant divergences (DIG-MILP PC1 \(W_1 \approx 1.47 - 1.69\)). ACM-MILP (IS) demonstrated intermediate similarity (PC1 \(W_1 \approx 0.89 - 1.01\)).

\begin{figure}
    \centering
    \includegraphics[width=1.0 \linewidth]{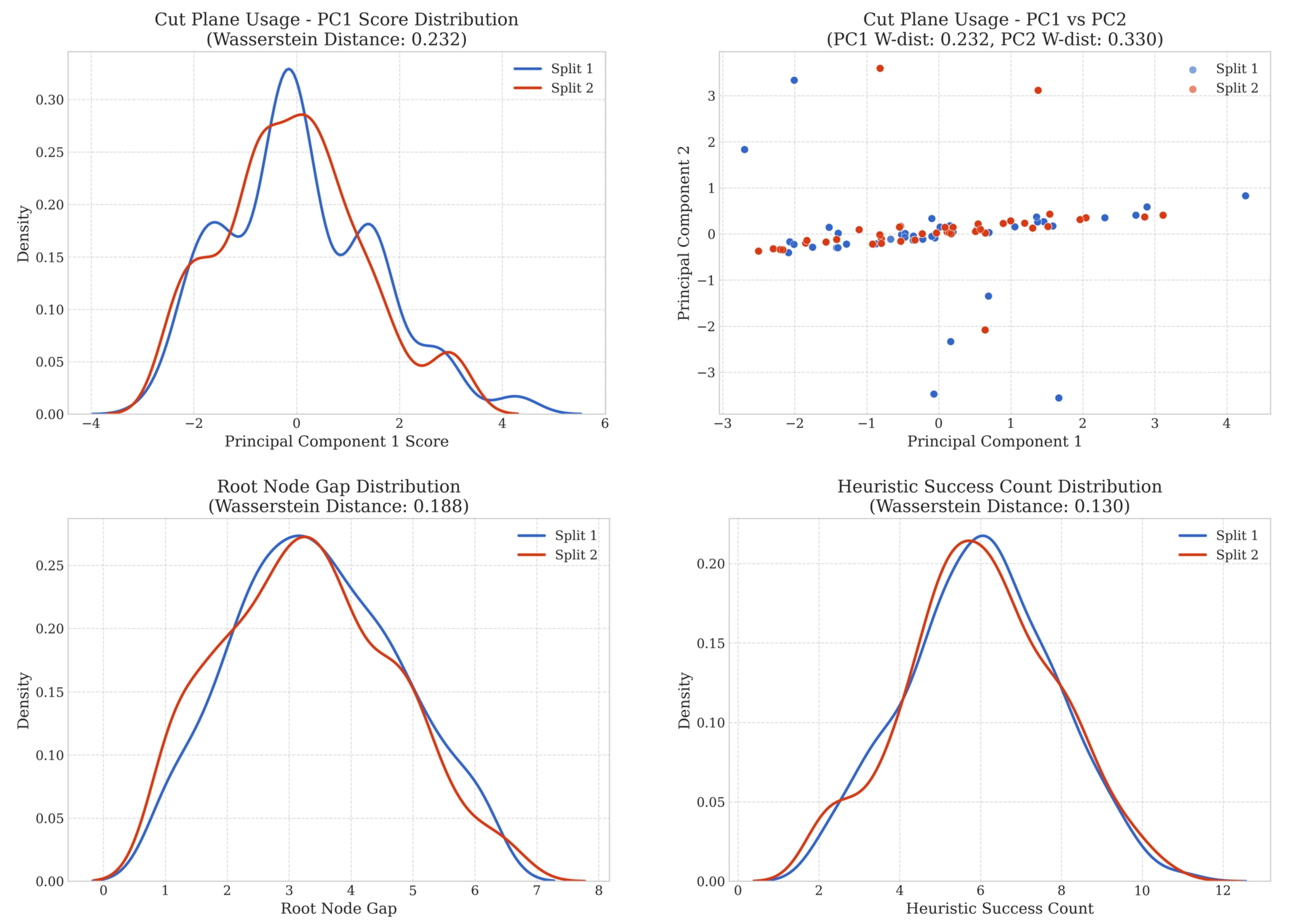}
    \caption{\textbf{Internal Cut-plane Comparison of IS (Demonstration of Identical Distribution)}}
    \label{fig:is_sif}
\end{figure}

%% file: experiments.tex
\section{Experiments}

\subsection{Efficiency Test on Super Hard Instances} \label{ml4co}

To validate the practical feasibility and robustness of our solver-internal feature analysis method, we designed an efficiency test.  
The core goal of this experiment is to demonstrate that even under strict computational constraints (i.e., tight time limits) and when facing a well-known set of hard instances, our method can still stably and efficiently extract meaningful feature distributions.  
This supports our key hypothesis that obtaining such deep behavioral features does not require solving instances to optimality, thereby making the evaluation both low-cost and highly efficient.

\begin{table}[ht]

    \centering

    \caption{\textbf{W-1 distances of solver-internal features across item placement halves}}

    \label{tab:efficiency-hard-wrapped}

    \begin{tabular}{l c}

    \toprule

    \textbf{Feature} & \textbf{W-1 Dist.} \\

    \midrule

    Root Node Gap & 0.7613 \\

    Heuristic Success Count & 0.7320 \\

    Cut Plane Usage & \\

    \quad -- PC1 & 0.0917 \\

    \quad -- PC2 & 0.0870 \\

    \quad -- PC3 & 0.0904 \\

    \bottomrule

    \end{tabular}

\end{table}

\textbf{Dataset Selection}: We used the item placement dataset from the ML4CO competition, which is recognized for its high solving difficulty. A total of 1000 instances were selected for testing.

\textbf{Computational Budget Constraint}: To simulate scenarios with limited computational resources, we imposed a strict time limit of \textbf{120 seconds} for the Gurobi solver on each instance.

\textbf{Validation Method}: We employed a split-half validation approach to evaluate the stability of the extracted features. The 1000 solver logs were randomly divided into two halves of 500 instances each. We then independently extracted solver-internal features—including Root Node Gap, Heuristic Success Count, and Cut Plane Usage—from both halves and computed the distributional similarity between them. Strong stability and reliability of the extraction method are indicated if these random subsets, originating from the same underlying distribution, exhibit highly similar feature distributions (i.e., a low Wasserstein distance).

The split-half validation demonstrates remarkable stability in the extracted solver-internal features, even under the strict 120-second time limit. As shown in Table~\ref{tab:efficiency-hard-wrapped}, the Cut Plane Usage metric exhibits exceptionally low 1-Wasserstein distances for its top three principal components (PC1: 0.0917, PC2: 0.0870, and PC3: 0.0904). These near-zero values indicate that the solver's strategic application of cuts—a crucial early-stage behavior—is highly consistent across the two random data subsets. Similarly, the Root Node Gap (0.7613) and Heuristic Success Count (0.7320) show strong distributional similarity, confirming that a stable signal for these metrics can also be captured long before the solver reaches optimality. Collectively, these results validate that a truncated solving process is sufficient to extract a reliable and consistent "fingerprint" of solver behavior, confirming the low-cost and efficient nature of our methodology.

\textbf{Conclusion:} This experiment strongly demonstrates the practical value of our framework: \circled{\scriptsize 1}~\textbf{High Efficiency and Low Cost}: The results confirm that a truncated solving process (120-second timeout) is sufficient to extract stable and representative solver-internal feature distributions. This substantially reduces the time and computational cost of evaluating large-scale or extremely hard instance sets. \circled{\scriptsize 2}~\textbf{Feature Stability}: The very small Wasserstein distances observed for Root Node Gap and Cut Plane Usage indicate that these early-stage solver behavior features are highly stable. \circled{\scriptsize 3}~\textbf{Practical Feasibility}: The success of this test shows that the GenBench-MILP framework is not only theoretically sound but also practically applicable for evaluating MILP instances that are difficult to solve to optimality within a reasonable time frame and that reflect real-world complexity.

\subsection{Cross-Solver Comparison and Stability Analysis} \label{crosssolver}

To assess the generality of the GenBench-MILP framework and probe the internal stability of different solvers, we extended our evaluation to include the open-source solvers SCIP and HiGHS. We conducted the same IS split-half validation procedure, using Gurobi as the stable baseline for comparison. Table~\ref{tab:cross-solver-wdist} summarizes the key results.
\vspace{-6pt}

\begin{table}[ht]
    \centering
    \caption{\textbf{1-Wasserstein distances across solvers on IS split-half validation}}
    \label{tab:cross-solver-wdist}
    \resizebox{\linewidth}{!}{%
    \begin{tabular}{lccc}
    \hline
    \textbf{Feature} & \textbf{Gurobi (Baseline)} & \textbf{SCIP} & \textbf{HiGHS} \\
    \hline
    Root Node Gap & \cellcolor{orange!30}\textbf{0.1884} & 1.7530 & \cellcolor{gray!20}26.7297 \\
    Heuristic (PC1) & \cellcolor{orange!30}\textbf{0.1300} & 0.3190 & 0.2368 \\
    Cut Plane (PC1) & 0.2318 & 0.7000 & \cellcolor{orange!30}\textbf{0.0612} \\
    \hline
    \end{tabular}%
    }
\end{table}

The data reveals a complex and multi-faceted picture of solver stability.
First, examining the \textbf{Root Node Gap}, we observe dramatic differences. Gurobi exhibits exceptional stability with a Wasserstein distance of just \textbf{0.1884}. In stark contrast, HiGHS shows a distance of 26.7297, over \textbf{140 times} larger than Gurobi's, signifying severe instability in its root node analysis. This suggests that the initial problem assessment and relaxation strategies in HiGHS are profoundly sensitive to small changes in problem structure.
Second, the results for the primary principal component of \textbf{Heuristic} features show a similar trend, though less pronounced. Gurobi remains the most stable (0.1300), while SCIP and HiGHS are approximately \textbf{2.5x} and \textbf{1.8x} less stable, respectively. This implies Gurobi's primal heuristic strategies are more robust.
Interestingly, the analysis of \textbf{Cut Plane} features presents a contrasting result. Here, HiGHS is by far the most stable solver (0.0612), with a consistency nearly 4 times greater than Gurobi's (0.2318). This suggests that HiGHS may employ a more deterministic or conservative cut generation strategy, whereas Gurobi and SCIP might use more adaptive, opportunistic approaches that are inherently more variable.

From this detailed analysis, we draw two main conclusions. First, the successful application to SCIP and HiGHS validates that \textit{\textbf{the GenBench-MILP framework is indeed solver-agnostic and generalizable}}. Second, and more importantly, the quality of instance evaluation is intrinsically linked to solver stability, which is not monolithic but \textit{highly feature-dependent}. Gurobi's superior stability in the critical root-node phase makes it the most reliable choice for a baseline. The extreme volatility of the root node gap in HiGHS, for instance, could obscure subtle computational similarities that a more stable solver like Gurobi can reliably detect.

In summary, this cross-solver comparison highlights a powerful secondary application of our framework. Beyond its primary purpose of instance evaluation, \textit{GenBench-MILP serves as an effective diagnostic tool for profiling the internal dynamics of MILP solvers}, offering quantitative insights into their stability and behavioral responses to perturbations across different strategic components.

%% file: discussion.tex
\section{Discussion} \label{discussion}

\noindent \textbf{Finding 1:}  \textit{\textbf{Superficial structural similarity is an unreliable predictor of computational behavior and difficulty in generated MILP instances.}} ACM-MILP (CA) instances showed high structural similarity but were exceptionally difficult to solve, with Root Node Gap discrepancies suggesting that structural metrics miss key complexity drivers or that models introduce subtle, difficulty-enhancing variations.

\noindent \textbf{Finding 2:}  \textit{\textbf{Simple outcome metrics, such as solving time or branching nodes, do not adequately reveal the source of problem difficulty or underlying changes to instance structure.}} This is exemplified by ACM-MILP (IS) instances, which, despite some similar outcome metrics (e.g., solving time), exhibited vastly different heuristic behavior, indicating altered internal structures that impact solvability in ways not captured by these surface-level performance measures.

\noindent \textbf{Finding 3:} \textit{\textbf{The effectiveness of generative models is problem-dependent, linked to how well their modification strategies align with the core structural determinants of hardness for specific problem types.}} G2MILP's difficulty in generating realistic IS instances is likely due to IS problem hardness being rooted in global graph topology \citep{tarjan1977finding, west2001introduction}, which its local edge constraint modifications struggle to preserve. G2MILP's better performance on SC problems suggests its direct manipulation of constraint configurations \citep{balas1980} more effectively captures their structural hardness.

%% file: conclusion.tex
\section{Conclusion} 

We introduce GenBench-MILP
, a comprehensive benchmark framework designed to address the critical shortcomings of traditional evaluation methods for generated MILP instances. Our findings reveal that metrics based on superficial structural similarity or simple outcomes often fail to capture an instance's true computational nature and difficulty. GenBench-MILP facilitates a more accurate and holistic assessment by shifting the focus towards solver-perceived difficulty and the practical utility of instances in downstream applications. By enabling this multifaceted methodology, our framework will be instrumental in guiding the development of higher-quality instance generators and fostering impactful research within the field of mathematical optimization.

%% file: impact_statement.tex
\section*{Impact Statement}

This paper presents work whose goal is to advance the field of Machine Learning. There are many potential societal consequences of our work, none which we feel must be specifically highlighted here.

%% file: appendix.tex
\section{Datasets} \label{datasets}

\subsection{Ecole Synthesized Instances} \label{sec:datasets}

To facilitate various experiments, including model training and evaluation across different metrics, 4 synthetic datasets of MILP instances was generated. This process utilized the \textbf{Ecole} library \citep{prouvost2020ecole}, a platform designed for machine learning research in combinatorial optimization.

\textbf{Parameter Randomization:} Parameters like the number of constraints, variables, and density are randomly sampled for each instance within the ranges specified in the configuration file. This ensures diversity in the generated dataset. The details of synthesizing parameters are in Table~\ref{tab:synthesis_params}.
\vspace{6pt}

\begin{table}[htbp]
\centering
\caption{Parameter Settings for MILP Instance Synthesis by Problem Type}
\label{tab:synthesis_params}
\small 
\begin{tabular}{@{}lll@{}}
\toprule
Problem Type                     & Parameter                    & Value / Range          \\ \midrule
Set Cover (SC)                   & Constraints Range            & [200, 800]             \\
                                 & Variables Range              & [400, 1600]            \\
                                 & Density Range                & [0.05, 0.2]            \\ \midrule
Capacitated Facility             & Ratio (Facilities/Customers) & 0.5                    \\
Location (CFL)* & Constraints Range            & [50, 150]              \\
                                 & Variables Range              & [500, 5000]            \\
                                 & Density Range                & [0.01, 0.3]            \\ \midrule
Combinatorial Auction (CA)** & Number of Items (n\_items)   & Auto-calculated (null) \\
                                 & Value Range (min/max)        & [1, 100]               \\
                                 & Constraints Range            & [50, 200]              \\
                                 & Variables Range              & [80, 600]              \\
                                 & Density Range                & [0.02, 0.1]            \\ \midrule
Independent Set (IS)*** & Variables (Nodes) Range      & [480, 520]             \\
                                 & Density (Edge Prob.) Range   & [0.01, 0.015]          \\
                                 & Constraints Range            & N/A (Derived)          \\
\bottomrule
\multicolumn{3}{p{0.9\textwidth}}{\footnotesize *For CFL, the input `constraints` and `variables` ranges influence the number of customers and facilities, which in turn determine the actual model size. The `density` range specified might not be directly used by the `ecole` generator for CFL.} \\
\multicolumn{3}{p{0.9\textwidth}}{\footnotesize **For CA, the input `constraints`, `variables`, and `density` ranges influence the number of bids and potentially items (if `n\_items` is null), which determine the actual model size.} \\
\multicolumn{3}{p{0.9\textwidth}}{\footnotesize ***For IS, the `variables` range directly corresponds to the number of nodes. The `constraints` range is implicitly determined by the number of nodes and the edge probability (density) and is not directly set via `min/max\_constraints` in the config.} \\
\end{tabular}
\end{table}

\subsection{Compatibility Testing with Public Datasets (NIPS ML4CO)}

In addition to utilizing the synthetically generated dataset described above, the compatibility and robustness of the various analysis pipelines and tools developed (e.g., for feature extraction, solver log parsing, model evaluation) were verified using publicly available MILP benchmark instances. Specifically, datasets from the \textbf{ML4CO Competition} \citep{gasse2022ml4co} were employed for testing purposes. Successfully processing and running analyses on these standard benchmarks demonstrated the broader applicability and compatibility of the developed experimental framework beyond the specific synthetically generated instances. More results are in \hyperlink{ml4co}{Appendix L.3}

\section{More Related Work} \label{more_related_work}

The generation of high-quality MILP instances is crucial for developing, testing, and tuning both traditional solvers and modern ML approaches for combinatorial optimization. However, the scarcity of diverse, representative real-world instances, often due to proprietary constraints or collection difficulties, presents a significant bottleneck. This has spurred research into synthetic instance generation techniques. Recently, deep learning (DL) has emerged as a promising direction for MILP instance generation, aiming to automatically learn complex data distributions and structural features from existing instances without requiring explicit, expert-designed formulations. Several DL-based frameworks have been proposed:

    \textbf{VAE-based Approaches:} G2MILP \citep{geng2023} introduced the first DL-based framework, using a masked Variational Autoencoder (VAE) paradigm. It iteratively corrupts and replaces constraint nodes in the bipartite graph representation. While capable of generating instances structurally similar to the training data, G2MILP does not explicitly guarantee the feasibility or boundedness of the generated instances, potentially leading to unusable samples for certain downstream tasks. DIG-MILP \citep{wang2023}, also VAE-based, addresses the feasibility and boundedness guarantee by leveraging MILP duality theories and sampling from a space including feasible solutions for both primal and dual formats. However, this can come at the cost of structural similarity compared to the original data. ACM-MILP \citep{guo2024} refines the VAE approach by incorporating adaptive constraint modification. It uses probability estimation in the latent space to select instance-specific constraints for modification, preserving core characteristics. It also groups strongly related constraints via community detection for collective modification, aiming to maintain constraint interrelations and improve hardness preservation.

    \textbf{Diffusion-based Approaches:} Recognizing the power of diffusion models in generation tasks, MILP-FBGen \citep{zhang2024} proposed a diffusion-based framework. It uses a structure-preserving module to modify constraint/variable vectors and a feasibility/boundedness-constrained sampling module for the right-hand side and objective coefficients, aiming to ensure both structural similarity and feasibility/boundedness. Another diffusion-based approach focuses specifically on generating \textit{feasible solutions} (rather than full instances) using guided diffusion. It employs contrastive learning to embed instances and solutions and uses a diffusion model conditioned on instance embeddings to generate solution embeddings, guided by constraints and objectives during sampling.

    \textbf{Structure-Focused Approaches:} Some recent works explicitly target specific structural properties. MILP-StuDio \citep{liu2024} focuses on preserving and manipulating \textit{block structures} commonly found in CCMs of real-world MILPs. It decomposes instances into block units, builds a library of these units, and uses operators (reduction, mix-up, expansion) to construct new, potentially scalable instances while maintaining structural integrity and properties like hardness and feasibility. MILPGen \citep{yang_mipgen_2023}also aims at scalability, using node splitting/merging on the bipartite graph representation to decompose instances into tree structures, which are then concatenated to build larger problems.


These learning-based methods offer the potential to generate diverse and realistic MILP instances automatically, addressing the data scarcity issue. The generated instances have shown promise in downstream tasks like solver hyperparameter tuning, data augmentation for ML models predicting optimal values or guiding solvers, and potentially constructing harder benchmarks.

\section{Parameter Summary} \label{summary}

See the parameters of our methods in Table~\ref{tab:exp_params_all}.

{\small
\begin{longtable}{@{}lllp{0.25\textwidth}@{}}
\caption{Summary of Key Experimental Parameters.}
\label{tab:exp_params_all} \\

\toprule
\textbf{Category} & \textbf{Parameter Name} & \textbf{Value} & \textbf{Description} \\
\midrule
\endfirsthead

\caption{Summary of Key Experimental Parameters (Continued).} \\
\toprule
\textbf{Category} & \textbf{Parameter Name} & \textbf{Value} & \textbf{Description} \\
\midrule
\endhead

\bottomrule
\endfoot
\bottomrule
\endlastfoot

\textbf{Feasibility Ratio}
 & Gurobi \texttt{time\_limit} & 300 s & Maximum Gurobi solve time per instance for feasibility/boundedness check. \\
\midrule

\multirow{2}{*}{\textbf{Structural Similarity}}
 & Feature Set & 11 features & Set of graph-based structural features extracted (e.g., density, degree stats). \\
 & Comparison Metric & JSD-based & Jensen-Shannon Divergence ($1 - JSD / \log(2)$) averaged. \\
\midrule

\multirow{3}{*}{\textbf{Branching Nodes}}
 & Gurobi \texttt{time\_limit} & 300 s & Maximum Gurobi solve time per instance. \\
 & Gurobi \texttt{threads} & 1 & Threads per solve (to enhance determinism). \\
 & Metric & Relative Error & $|(\sum N_{gen} - \sum N_{train}) / \sum N_{train}| \times 100\%$. \\
\midrule

\multirow{3}{*}{\textbf{Solving Time Gap}}
 & Gurobi \texttt{time\_limit} & 3600 s & Maximum Gurobi solve time per instance. \\
 & Gurobi \texttt{threads} & 1 & Threads per solve. \\
 & Metric & Relative Difference & $|(\text{mean\_t}_{gen} - \text{mean\_t}_{train}) / \text{mean\_t}_{train}| \times 100\%$. \\
\midrule

\multirow{6}{*}{\textbf{Hyperparameter Tuning}}
 & Tuning Framework & SMAC3 & Automated algorithm configuration tool used. \\
 & \texttt{n\_trials} & 50 & SMAC3 evaluation budget. \\
 & Objective & Mean Solve Time & Minimize mean wall-clock time on the tuning set. \\
 & Gurobi \texttt{time\_limit} & 60 s & Gurobi time limit per solve during tuning/evaluation. \\
 & Gurobi \texttt{threads} & 1 & Gurobi threads per solve during tuning/evaluation. \\
 & Tuned Parameters & Gurobi Controls & 8 parameters including \texttt{Heuristics}, \texttt{MIPFocus}, etc. \\
\midrule

\multirow{10}{*}{\textbf{Initial Basis Prediction}}
 & \multicolumn{3}{l}{\textit{GNN Model Parameters}} \\
 & \texttt{num\_layers} & 3 & Number of GNN message passing layers. \\
 & \texttt{hidden\_dim} & 128 & Dimensionality of hidden layers in the GNN. \\
 & \texttt{dropout} & 0.1 & Dropout rate used during training. \\
 \addlinespace
 & \multicolumn{3}{l}{\textit{GNN Training Parameters}} \\
 & Loss Function & Weighted CE & Weighted Cross-Entropy + Label Smoothing (0.1). \\
 \addlinespace
 & \multicolumn{3}{l}{\textit{Evaluation Parameters}} \\
 & Gurobi \texttt{time\_limit} & 600 s & Gurobi time limit per solve during evaluation. \\
 & Gurobi \texttt{threads} & 1 & Gurobi threads per solve during evaluation. \\
 & Basis Repair Threshold & 1e-12 & Pivot threshold for basis repair stability check. \\
\midrule

\multirow{7}{*}{\textbf{Solver-Internal Features}}
 & \multicolumn{3}{l}{\textit{Data Generation Phase}} \\
 & Gurobi \texttt{time\_limit} & 100 s & Gurobi time limit for log generation. \\
 \addlinespace
 & \multicolumn{3}{l}{\textit{Comparison Phase}} \\
 & Root Gap Metric & 1-Wasserstein & Metric to compare scalar distributions. \\
 & Cut Plane Metric & PCA + 1-W. & Method for comparing high-dim cut vectors. \\
 & PCA Components & 3 & Number of components retained for cut analysis. \\
 & PCA Scaling & Standard & Data scaling method (Z-score) before PCA. \\

\end{longtable}
}

\section{Experiment Environment} \label{envir}

All experiments were conducted on a Linux-based system equipped with an AMD EPYC 9754 128-Core Processor with a typical operating frequency around 3.1 GHz and supports 256 threads (128 cores with 2 threads per core). The system has 256 MiB of L3 cache and 251 GiB of system RAM.

For GPU-accelerated computations, a NVIDIA GeForce RTX 3090 graphics card with 24 GiB of VRAM was utilized. The NVIDIA driver version was 535.171.04, supporting CUDA up to version 12.2. The CUDA Toolkit version used for development and compilation was 11.8 (V11.8.89).

The operating system was Ubuntu 20.04.6 LTS. Key software includes Python 3.8.20, Gurobi Optimizer version 11.0.1.

\section{More Preliminaries} \label{more preli}

\paragraph{Simplex Method Basics} \quad The \textbf{Simplex Method} \citep{dantzig1963linear} is an algorithm primarily designed for solving Linear Programming (LP) problems, such as the LP relaxation defined above. While it can handle various forms, it often operates by converting the problem to a standard form like $\min \{ \mathbf{c'}^\top \mathbf{y} \mid \mathbf{A'} \mathbf{y} = \mathbf{b'}, \mathbf{y} \geq 0 \}$ (where $\mathbf{y}$ might include original and slack/surplus variables). It explores the vertices (Basic Feasible Solutions, BFS) of the feasible polyhedron $\mathcal{P}' = \{\mathbf{y} \mid \mathbf{A'} \mathbf{y} = \mathbf{b'}, \mathbf{y} \geq 0 \}$. It iteratively moves between adjacent BFS by exchanging one variable in the current basis (a set of variables determining the BFS) with a non-basic variable, guided by criteria like reduced costs to ensure monotonic improvement of the objective function value, until an optimal BFS is found or unboundedness is detected.

\paragraph{Heuristics} \quad Within MILP solvers, \textbf{heuristics} \citep{berthold2012primal} are algorithms designed to rapidly find a feasible integer solution $\hat{\mathbf{x}} \in \mathcal{F}_{MILP} = \{\mathbf{x} \mid \mathbf{A} \mathbf{x} \leq \mathbf{b}, x_i \in \mathbb{Z} \; \forall i \in \mathcal{I}\}$, whose objective value $\mathbf{\tilde{c}}^\top \hat{\mathbf{x}}$ is hopefully close to the true optimum $z_{MILP}$. They do not guarantee optimality. Their main purpose is to quickly establish strong primal bounds (updating the incumbent $z_{incumbent} = \min(z_{incumbent}, \mathbf{\tilde{c}}^\top \hat{\mathbf{x}})$). In branch-and-bound, these incumbents enable pruning of search nodes $j$ where the local lower bound $z_{bound}^{(j)}$ satisfies $z_{bound}^{(j)} \geq z_{incumbent}$.

\paragraph{Cutting Planes (Cuts)} \quad \textbf{Cutting planes} \citep{conforti2014integer} are linear inequalities, $\alpha^T \mathbf{x} \leq \beta$, that are valid for all feasible integer solutions, i.e., they hold for all $\mathbf{x} \in Conv(\mathcal{F}_{MILP})$ where $\mathcal{F}_{MILP} = \{\mathbf{x} \mid \mathbf{A} \mathbf{x} \leq \mathbf{b}, x_i \in \mathbb{Z} \; \forall i \in \mathcal{I}\}$. However, they are chosen such that they are violated by the optimal solution $\mathbf{x}^*_{LP}$ of the current LP relaxation (i.e., $\alpha^T \mathbf{x}^*_{LP} > \beta$). Adding such cuts $\mathcal{C}$ to the LP relaxation feasible set $\mathcal{P} = \{\mathbf{x} \mid \mathbf{A} \mathbf{x} \leq \mathbf{b}\}$ yields a tighter relaxation $\mathcal{P}' = \mathcal{P} \cap \{\mathbf{x} \mid \alpha^T \mathbf{x} \leq \beta$  for all, $ \alpha^T \mathbf{x} \leq \beta \in \mathcal{C}\}$. The goal is to improve the lower bound derived from the relaxation, $\min_{\mathbf{x} \in \mathcal{P}'} \mathbf{\tilde{c}}^\top \mathbf{x} \geq z_{LP}$, thereby bringing it closer to the true MILP optimum $z_{MILP}$ and potentially reducing the search space.

\section{Feasibility}

Below is the result of feasibility test.

\vspace{5pt}
\begin{table}[htbp]
\centering
\caption{Feasibility and Boundedness Ratios Across All Problem Types. Each configuration was tested on 1000 instances.}
\label{tab:feasibility_all}
\begin{tabular}{llcc}
\toprule
\textbf{Problem} & \textbf{Model} & \textbf{Ratio ($\eta$)} & \textbf{Feasible (\%)} \\
\midrule
\multirow{7}{*}{Independent Set} & Original Set & N/A & 100.00 \\
\cmidrule{2-4}
 & \multirow{3}{*}{G2MILP} & 0.01 & 100.00 \\
 & & 0.05 & 100.00 \\
 & & 0.10 & 93.40 \\
\cmidrule{2-4}
 & \multirow{3}{*}{ACM-MILP} & 0.05 & 100.00 \\
 & & 0.10 & 100.00 \\
 & & 0.20 & 100.00 \\
\midrule
\multirow{7}{*}{Combinatorial Auction} & Original Set & N/A & 100.00 \\
\cmidrule{2-4}
 & \multirow{3}{*}{ACM-MILP} & 0.05 & 100.00 \\
 & & 0.10 & 100.00 \\
 & & 0.20 & 100.00 \\
\cmidrule{2-4}
 & \multirow{3}{*}{DIG-MILP} & 0.01 & 100.00 \\
 & & 0.05 & 100.00 \\
 & & 0.10 & 100.00 \\
\midrule
\multirow{4}{*}{Set Cover} & Original Set & N/A & 100.00 \\
\cmidrule{2-4}
 & \multirow{3}{*}{G2MILP} & 0.01 & 100.00 \\
 & & 0.05 & 100.00 \\
 & & 0.10 & 100.00 \\
\bottomrule
\end{tabular}
\end{table}

\section{Structural Similarity}

The experiment's objective is to quantitatively compare the structural similarity between two sets of MILP instances. This is achieved by extracting graph-based structural features from each instance and then calculating the Jensen-Shannon Divergence (JSD) to measure the difference between the feature distributions of the two sets.

\subsection{Methodology}

\textbf{1. Feature Extraction:} Each MILP instance is converted into a bipartite graph. From this graph, a vector of structural features is extracted, capturing characteristics related to matrix sparsity, variable and constraint connectivity, coefficient statistics, and other graph-theoretic properties. This process is computationally parallelized for efficiency. The specific features are grouped into two categories and detailed in Table~\ref{tab:structural_features}.

\begin{table}[htbp]
\centering
\caption{Structural Features Used for Similarity Analysis}
\label{tab:structural_features}
\begin{tabularx}{\linewidth}{@{} l l l >{\raggedright\arraybackslash}X @{}}
\toprule
\textbf{Category} & \textbf{Feature} & \textbf{Metric(s)} & \textbf{Description} \\
\midrule
\multirow{4}{*}{\begin{tabular}[c]{@{}l@{}}Basic Graph\\ Features\end{tabular}} 
 & Coefficient Density & \texttt{coef\_dens} & Reflects the sparsity of the constraint matrix. \\
 & Variable Node Degree & \texttt{var\_degree\_mean/std} & Describes the occurrence patterns of variables in constraints. \\
 & Constraint Node Degree & \texttt{cons\_degree\_mean/std} & Reflects the number of variables involved in each constraint. \\
 & Coefficient Statistics & \texttt{lhs/rhs\_mean/std} & Captures the numerical distribution of constraint terms. \\
\midrule
\multirow{2}{*}{\begin{tabular}[c]{@{}l@{}}Advanced\\ Topological\end{tabular}} 
 & Clustering Coefficient & \texttt{clustering} & Measures the local connection density between variables. \\
 & Modularity & \texttt{modularity} & Assesses the degree of modularity in the problem structure. \\
\bottomrule
\end{tabularx}
\end{table}

\textbf{2. Per-Feature Similarity:} For each feature, the Jensen-Shannon Divergence is used to compare its distribution across the two sets of instances. This divergence value is then normalized to a similarity score ranging from 0 (maximal divergence) to 1 (identical distributions).

\textbf{3. Overall Similarity Score:} A final, overall similarity score is calculated by taking the arithmetic mean of all the individual feature similarity scores. This single metric represents the aggregate structural similarity between the two instance sets.

\section{Solving Time Gap}

This experiment assesses the similarity in computational hardness between a 'training' set and a 'generated' set of MILP instances. The metric is the relative percentage difference in their average solve times using the Gurobi optimizer.

\textbf{Methodology:} Each instance from both sets is solved under identical Gurobi configurations to ensure a fair comparison. The key parameters are:
\begin{itemize}
    \item \textbf{Time Limit:} 300 seconds per instance.
    \item \textbf{Threads:} 256 per instance.
\end{itemize}

The average solve time is computed for each set. The final metric is calculated using the formula:
\[
\text{Relative Time Gap} = \frac{|\text{generated\_mean\_time} - \text{training\_mean\_time}|}{\max(\text{training\_mean\_time}, 10^{-10})} \times 100\%
\]
A lower percentage signifies a greater similarity in computational hardness between the two instance sets.

\subsection{Results}
The results are in Table~\ref{tab:solving_time_gap}. We have omitted the G2MILP-generated Independent Set (IS) instances from this presentation. The rationale for this decision lies in the substantial surge in their computational complexity compared to the baseline instances, leading to a prohibitive increase in solution times by a factor of several thousands.

\begin{table}[h]
  \centering
  \caption{\textbf{Solving Time Gap Comparison}. The table shows the percentage difference in average solving time between instances generated by different sources (with varying mask ratios $\eta$) and the original training set instances. \textbf{Original} refers to the baseline training set. G2MILP instances for IS reached the 300s time limit.}
  \label{tab:solving_time_gap}
  \setlength{\tabcolsep}{5pt}
  \renewcommand{\arraystretch}{1.0}
  \begin{tabular}{@{}llrrr@{}}
    \toprule
    Dataset & Source & $\eta$ & Avg Time (s) & Solving Time Gap (\%) \\
    \midrule
    \textbf{SC} & Original & --- & 0.2644 & --- \\
    \cmidrule(l){2-5}
                & G2MILP   & 0.10 & 0.3002 & \textbf{13.54} \\
                &          & 0.05 & 0.2945 & \textbf{11.38} \\
                &          & 0.01 & 0.3237 & 22.43 \\
    \midrule
    \textbf{CA} & Original & --- & 0.1020 & --- \\
    \cmidrule(l){2-5}
                & ACM-MILP & 0.20 & 2.6449 & 2493.04 \\
                &          & 0.10 & 3.1654 & 3003.33 \\
                &          & 0.05 & 3.4461 & \textbf{3278.53} \\
    \cmidrule(l){2-5}
                & DIG-MILP & 0.10 & 0.1312 & 28.63 \\
                &          & 0.05 & 0.1225 & \textbf{20.10} \\
                &          & 0.01 & 0.1237 & 21.27 \\
    \midrule
    \textbf{IS} & Original & --- & 0.3374 & --- \\
    \cmidrule(l){2-5}
                & ACM-MILP & 0.20 & 0.1722 & 48.96 \\
                &          & 0.10 & 0.1774 & 47.42 \\
                &          & 0.05 & 0.1788 & 47.01 \\
    \cmidrule(l){2-5}
                & G2MILP   & 0.10 & 300.00 & 88815.23 \\
                &          & 0.05 & 300.00 & 88815.23 \\
                &          & 0.01 & 300.00 & 88815.23 \\
    \bottomrule
  \end{tabular}
\end{table}

\section{Hyperparameter Tuning for Gurobi}

The objective of this project is to automatically tune a key set of Gurobi solver hyperparameters using the SMAC3 framework. The goal is to find a parameter configuration that \textbf{minimizes the average wall-clock solve time} for a specific collection of MILP instances, designated as the \textbf{tuning set}. Subsequently, the performance of this optimized configuration is compared against Gurobi's default settings on a separate, unseen \textbf{test set} to evaluate its generalization capability.

\subsection{Methodology}

The process is divided into five phases to ensure robust optimization and an unbiased evaluation.

\paragraph{Phase 1: Inputs and Setup}
The process utilizes two independent sets of MILP instances: a \textbf{tuning set}, which is used to train the SMAC3 model and guide its search for optimal parameters, and a separate \textbf{test set}, which is held out exclusively for the final performance evaluation to ensure an objective assessment. The core tuning engine is \textbf{SMAC3}, which directly interfaces with the target solver, \textbf{Gurobi}.

\paragraph{Phase 2: Hyperparameter Space Definition}
We selected eight critical Gurobi hyperparameters for optimization. The parameter space includes one continuous parameter, \texttt{Heuristics}, defined on the range [0.0, 1.0]. The remaining seven parameters are categorical: \texttt{MIPFocus} \{0, 1, 2, 3\}, \texttt{VarBranch} \{-1, 0, 1, 2, 3\}, \texttt{BranchDir} \{-1, 0, 1\}, \texttt{Presolve} \{-1, 0, 1, 2\}, \texttt{PrePasses} \{-1, \ldots, 20\}, \texttt{Cuts} \{-1, 0, 1, 2, 3\}, and \texttt{Method} \{-1, \ldots, 5\}.

\paragraph{Phase 3: Automated Tuning with SMAC3}
SMAC3 performs an iterative optimization over a budget of \textbf{50 trials}. Each trial begins with SMAC3 suggesting a new hyperparameter configuration from its internal model. Gurobi then uses this configuration to solve all instances in the \textbf{tuning set}, with each solve constrained to a \textbf{60-second time limit} and using \textbf{256 threads}. The resulting \textbf{average solve time} across the set is calculated and returned to SMAC3 as a cost metric. This feedback allows SMAC3 to update its model and inform its choice for the next trial. Upon completion, SMAC3 reports the \textbf{best configuration} found, which is the one that yielded the lowest average solve time.

\paragraph{Phase 4: Performance Evaluation on the Test Set}
To assess generalization, we compare the \textbf{best configuration} found by SMAC3 against Gurobi's \textbf{default configuration} on the independent \textbf{test set}. Both configurations are run under identical computational conditions (60-second time limit, 256 threads) to ensure a fair and direct comparison of their performance.

\paragraph{Phase 5: Outputs and Analysis}
The final outputs synthesize the results into two key components. The primary metric is a comparison of the average solve times on the test set for the best versus default configurations, often expressed as a percentage improvement. The analysis also provides the \textbf{best configuration} itself, detailing the specific values for the eight optimized hyperparameters that yielded the top performance.

Table~\ref{tab:hyperparameter_comparison} shows the best Gurobi hyperparameter values found by SMAC3 for each experiment. The default Gurobi parameter values are: \texttt{Heuristics}=0.05, \texttt{MIPFocus}=0, \texttt{VarBranch}=-1, \texttt{BranchDir}=0, \texttt{Presolve}=-1, \texttt{PrePasses}=-1, \texttt{Cuts}=-1, \texttt{Method}=-1.
\vspace{10pt}

\begin{table}[htbp]
  \centering
  \caption{Best Gurobi hyperparameter values found for each experiment. {Column Abbreviations:} Heur.: Heuristics, Focus: MIPFocus, VarBr.: VarBranch, BrDir.: BranchDir, Pres.: Presolve, PreP.: PrePasses, Cuts: Cuts, Meth.: Method.}
  \label{tab:hyperparameter_comparison}
  \begin{tabular}{lrrrrrrrr}
    \toprule
    Experiment Name        & Heur. & Focus & VarBr. & BrDir. & Pres. & PreP. & Cuts & Meth. \\
    \midrule
    acmmilp\_mis\_0.1      & 0.189 & 2     & 1      & -1     & -1    & 10    & 0    & 5     \\
    acmmilp\_mis\_0.2      & 0.474 & 0     & 1      & 0      & -1    & 17    & 1    & 2     \\
    acmmilp\_mis\_0.05     & 0.500 & 0     & -1     & -1     & -1    & -1    & -1   & -1    \\
    ca                   & 0.186 & 1     & 1      & 1      & 0     & 1     & -1   & 4     \\
    digmilp\_ca\_0.1       & 0.186 & 1     & 1      & 1      & 0     & 1     & -1   & 4     \\
    digmilp\_ca\_0.05      & 0.186 & 1     & 1      & 1      & 0     & 1     & -1   & 4     \\
    digmilp\_ca\_0.01      & 0.186 & 1     & 1      & 1      & 0     & 1     & -1   & 4     \\
    mis                  & 0.498 & 0     & 1      & -1     & 1     & -1    & 1    & 3     \\
    setcover             & 0.189 & 2     & 1      & -1     & -1    & 10    & 0    & 5     \\
    g2milp\_setcover\_0.01 & 0.189 & 2     & 1      & -1     & -1    & 10    & 0    & 5     \\
    g2milp\_setcover\_0.05 & 0.189 & 2     & 1      & -1     & -1    & 10    & 0    & 5     \\
    \bottomrule
  \end{tabular}
\end{table}

\section{Initial Basis Prediction} \label{ibp}

Following \citep{fan2023}, this experiment utilizes Graph Neural Networks (GNNs) to predict a high-quality initial basis for the LP relaxation of MILP instances. The objective is to train a GNN on **an augmented dataset (combining original and generated instances)** and evaluate whether its predicted basis accelerates Gurobi compared to default initialization. Performance is measured by Gurobi runtime, node count, and iteration count on an unseen test set.

\subsection{Methodology \& Operations}

The methodology employs a six-phase approach: transforming MILP instances into graph representations, feature extraction, model training, and performance evaluation. This process ensures robust basis prediction while maintaining numerical stability.

\paragraph{Phase 1: Problem Description and Data Representation}

MILP instances are defined by constraint matrix , RHS vector , objective , and variable bounds . The prediction task identifies an optimal initial basis, defined by sets  and  (where ), ensuring the basis matrix is non-singular.

Instances are transformed into bipartite graphs .  contains variable and constraint nodes;  represents structural relationships where edge  exists if , weighted by .

\paragraph{Phase 2: Feature Engineering}

Node representations capture both local structure and global characteristics.
Variable node features (8 dimensions) include: objective coefficient , variable density , and similarity to slack bounds (e.g., ). Bounds are encoded via numerical values and binary flags (finite vs. infinite).

Constraint node features (8 dimensions) include: similarity to objective , constraint density , and similarities to variable bounds. Continuous features (excluding binary flags) undergo z-score standardization for training stability.

\paragraph{Phase 3: Graph Neural Network Model}

The hierarchical GNN architecture processes the bipartite structure. An initial MLP projects 8-dimensional inputs to a 128-dimensional hidden space. The core comprises  \texttt{BipartiteMessagePassing} layers with residual connections to aggregate neighborhood information.

Separate MLP heads output 3-dimensional logits for variables and constraints, corresponding to basis states: NonbasicAtLower, Basic, and NonbasicAtUpper. Knowledge masking enforces physical constraints by blocking impossible states based on bounds. Final probabilities are obtained via softmax on masked logits, utilizing dropout (0.1) for regularization.

\paragraph{Phase 4: Model Training}

A \texttt{GNNTrainer} trains the model using **a composite dataset where generated instances are added to the original baseline**. The training runs for up to 800 epochs with a batch size of 32. The loss function uses label smoothing (0.1) to improve generalization and dynamic class weights to handle class imbalance. Optimization employs Adam (learning rate 1e-3) with L2 weight decay (1e-4). Early stopping terminates training if validation loss stagnates for 50 epochs, preserving the best-performing weights.

\paragraph{Phase 5: Initial Basis Generation and Repair}

GNN predictions are converted into a numerically stable basis. Candidates are selected based on the highest predicted "Basic" probabilities. To ensure non-singularity, a repair process performs LU decomposition on the candidate matrix, iteratively replacing unstable columns. Finally, nonbasic variables are assigned states (NonbasicAtLower or NonbasicAtUpper) according to their predicted probability distributions.

\paragraph{Phase 6: Performance Evaluation}

The evaluation compares the GNN-predicted basis against Gurobi's default initialization on an independent test set. Metrics include runtime, node count, and iteration count, providing a comprehensive assessment of the algorithmic efficiency gains achieved by the learned initialization.

\section{Solver-Internal Features}

This experiment analyzes the behavioral patterns of the Gurobi solver to compare original and generated sets of MILP instances. The methodology involves a two-phase process: first, we gather detailed performance metrics by solving each instance, and second, we apply statistical techniques to quantify the similarity between the instance sets based on the solver's behavior.

\subsection*{Methodology}

\paragraph{Phase 1: Data Generation and Metric Extraction}
In the first phase, we solve every MILP instance from both the original and generated collections using Gurobi under identical, standardized settings. This controlled environment ensures that any observed differences in performance are attributable to the instances themselves, not the solver's configuration. We enable detailed logging to capture a comprehensive record of the solver's internal operations.

From these logs, we systematically parse and extract key operational metrics. These metrics include the \textbf{duality gap at the root node}, which indicates initial problem difficulty; the \textbf{success counts of various heuristics}, which measure the effectiveness of the solver's solution-finding strategies; and the \textbf{usage frequency of different cut plane types} (e.g., Gomory, Cover, MIR), which reveals the solver's approach to tightening the LP relaxation. This process transforms the raw log files into structured data, ready for quantitative analysis.

\paragraph{Phase 2: Comparative Statistical Analysis}
The second phase quantifies the similarity between the original and generated instance sets by comparing the distributions of the metrics collected in Phase 1. We primarily use the \textbf{1-Wasserstein distance} as a robust measure of distributional difference—a smaller distance implies greater similarity.

For one-dimensional metrics like the root node gap and heuristic success counts, we directly compare their distributions. For the multi-dimensional cut plane data, we first normalize the usage counts for each instance into proportions. We then apply \textbf{Principal Component Analysis (PCA)} to reduce the dimensionality of this data, focusing the comparison on the most significant patterns of solver behavior. The Wasserstein distance is then used to compare the instance sets along these principal components. The final output consists of these distance values, which serve as a quantitative score of behavioral similarity between the two datasets.

\vspace{10pt}

\begin{table}[htbp]
\centering
\begin{minipage}[t]{0.45\textwidth}
\centering
\caption{Root Node Gap Statistics Comparison (Dataset Halves)}
\label{tab:root_node_gap_stats_test}
\begin{tabular}{lrr}
\toprule
\textbf{Statistic} & \textbf{part1} & \textbf{part2} \\
\midrule
count                & 500         & 500        \\
mean                 & 3.4637           & 3.4194           \\
std                  & 1.4386           & 1.4514           \\
min                  & 0.9722           & 1.0417           \\
25\%                  & 2.2222           & 2.4094           \\
50\%                  & 3.4483           & 3.2076           \\
75\%                  & 4.6289           & 4.5045           \\
max                  & 6.1364           & 6.5909           \\
W-Dist. & 0.1884           & 0.1884           \\
\bottomrule
\end{tabular}
\end{minipage}

\hspace{.5cm}

\begin{minipage}[t]{0.45\textwidth}
\centering
\caption{Heuristic Success Count Statistics Comparison (Dataset Halves)}
\label{tab:heuristic_stats_test}
\begin{tabular}{lrr}
\toprule
\textbf{Statistic} & \textbf{part1} & \textbf{part2} \\
\midrule
count                & 500        & 500      \\
mean                 & 6.0380           & 6.0400           \\
std                  & 1.7879           & 1.9541           \\
min                  & 2.0000           & 2.0000           \\
25\%                  & 5.0000           & 4.0000           \\
50\%                  & 6.0000           & 6.0000           \\
75\%                  & 7.0000           & 7.0000           \\
max                  & 11.0000          & 11.0000          \\
W-Dist. & 0.1300           & 0.1300           \\
\bottomrule
\end{tabular}
\end{minipage}
\end{table}

\vspace{10pt}

\begin{table}[t!]
\centering
\caption{Cut Plane Usage PCA 1-Wasserstein Distances (Dataset Halves)}
\label{tab:cutplane_pca_dist_test}
\begin{tabular}{lr}
\toprule
\textbf{Principal\_Component} & \textbf{W-Dist.} \\
\midrule
PC1                   & 0.2318 \\
PC2                   & 0.3295 \\
PC3                   & 0.2280 \\
\bottomrule
\end{tabular}
\end{table}

\begin{table}[ht] 
\centering
\caption{Overall Gurobi Solver Metrics Across Datasets}
\label{tab:solver_metrics_overall}
\begin{tabular}{@{} l r r @{}}
\toprule
Dataset & Avg. Root Gap (\%) & Heuristic Successes \\
\midrule
IS (Raw)                     & 3.42 & 18883 \\
Combinatorial Auction (Raw)  & 2.47 & 2115  \\
Set Cover (Raw)              & 7.61 & 2702  \\
ACM-MILP CA ($\eta=0.1$)     & 7.49 & 2552  \\
ACM-MILP IS ($\eta=0.1$)     & 2.66 & 1746  \\
DIG-MILP CA ($\eta=0.05$)    & 1.09 & 2241  \\
G2MILP SetCover ($\eta=0.05$)& 7.75 & 2600  \\
\bottomrule
\end{tabular}
\end{table}

\begin{table}[htbp] 
\centering
\caption{Aggregated Cut Plane Counts (Sum over 1000 instances)}
\label{tab:solver_metrics_cuts_combined}
\resizebox{\textwidth}{!}{%
\begin{tabular}{@{} l rrrrrr rrrrrr @{}}
\toprule
\textbf{Dataset} & \textbf{Gomory} & \textbf{ZeroHalf} & \textbf{Clique} & \textbf{MIR} & \textbf{RLT} & \textbf{FlowCover} & \textbf{Cover} & \textbf{ModK} & \textbf{RelaxLift} & \textbf{InfProof} & \textbf{StrongCG} & \textbf{ImplBound} \\
\midrule
IS (Raw)                     & 1305  & 61590 & 89    & 58   & 43763 & 0   & 0  & 0  & 0 & 0 & 0  & 0  \\
Combinatorial Auction (Raw)  & 11183 & 7014  & 8941  & 98   & 81    & 0   & 3  & 24 & 0 & 0 & 6  & 0  \\
Set Cover (Raw)              & 1350  & 3473  & 80    & 5601 & 240   & 0   & 0  & 8  & 0 & 0 & 0  & 1  \\
ACM-MILP CA ($\eta=0.1$)     & 6524  & 3132  & 26779 & 25   & 2     & 229 & 86 & 3  & 4 & 3 & 0  & 0  \\
ACM-MILP IS ($\eta=0.1$)     & 1087  & 34032 & 33    & 13   & 22932 & 0   & 0  & 1  & 0 & 0 & 0  & 0  \\
DIG-MILP CA ($\eta=0.05$)    & 6024  & 5629  & 462   & 531  & 114   & 0   & 14 & 34 & 0 & 1 & 14 & 0  \\
G2MILP SetCover ($\eta=0.05$)& 1330  & 3228  & 31    & 5454 & 209   & 0   & 0  & 8  & 0 & 0 & 0  & 0  \\
\bottomrule
\end{tabular}%
}
\end{table}

\subsection{Visualization} \label{validation}

\begin{figure}
    \centering
    \includegraphics[width=1\linewidth]{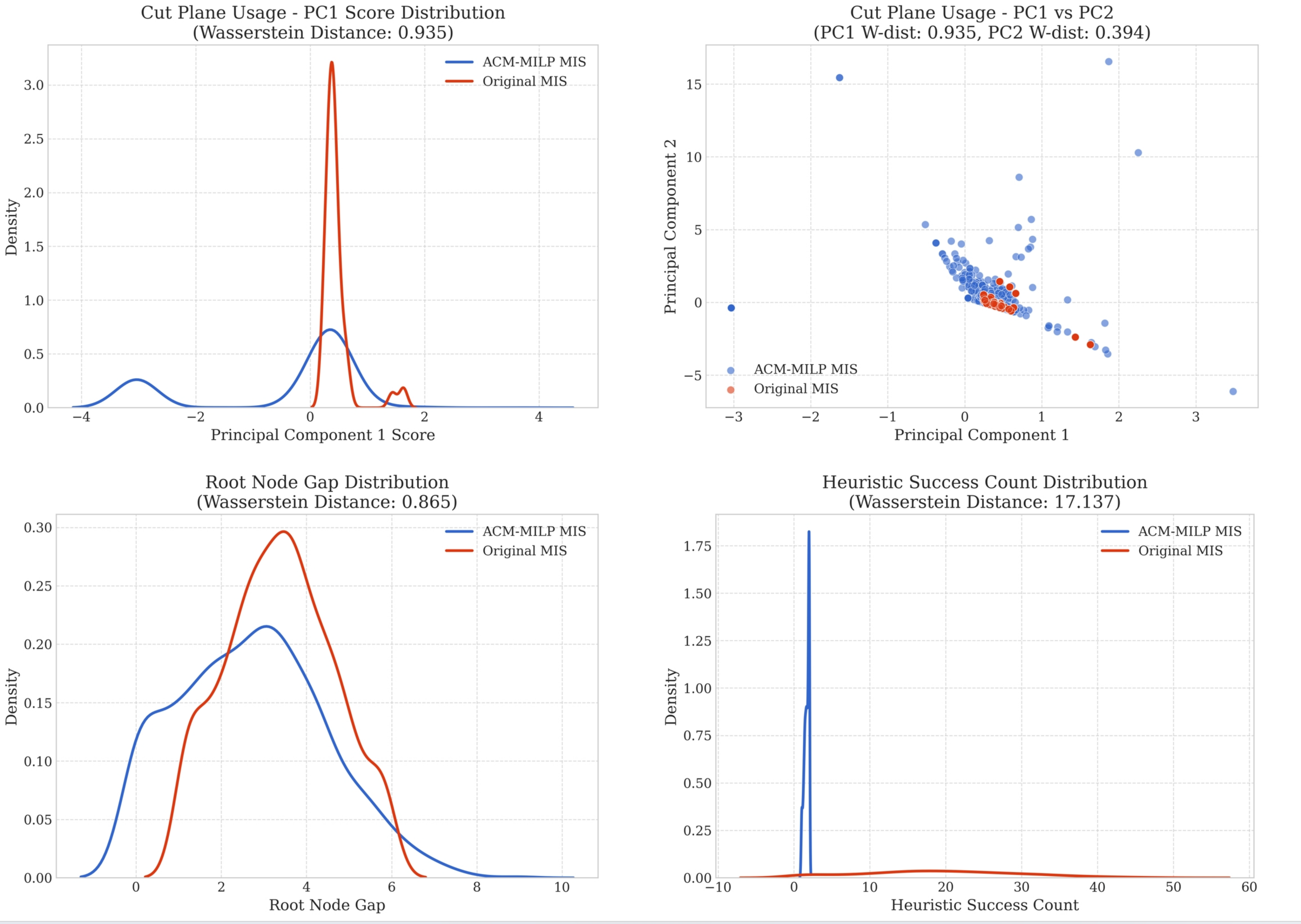}
    \caption{ACM-MILP IS at $\eta = 0.1$ compared with Original Dataset}
    \label{fig:acm_mis_0.1_sif}
\end{figure}

\begin{figure}
    \centering
    \includegraphics[width=1\linewidth]{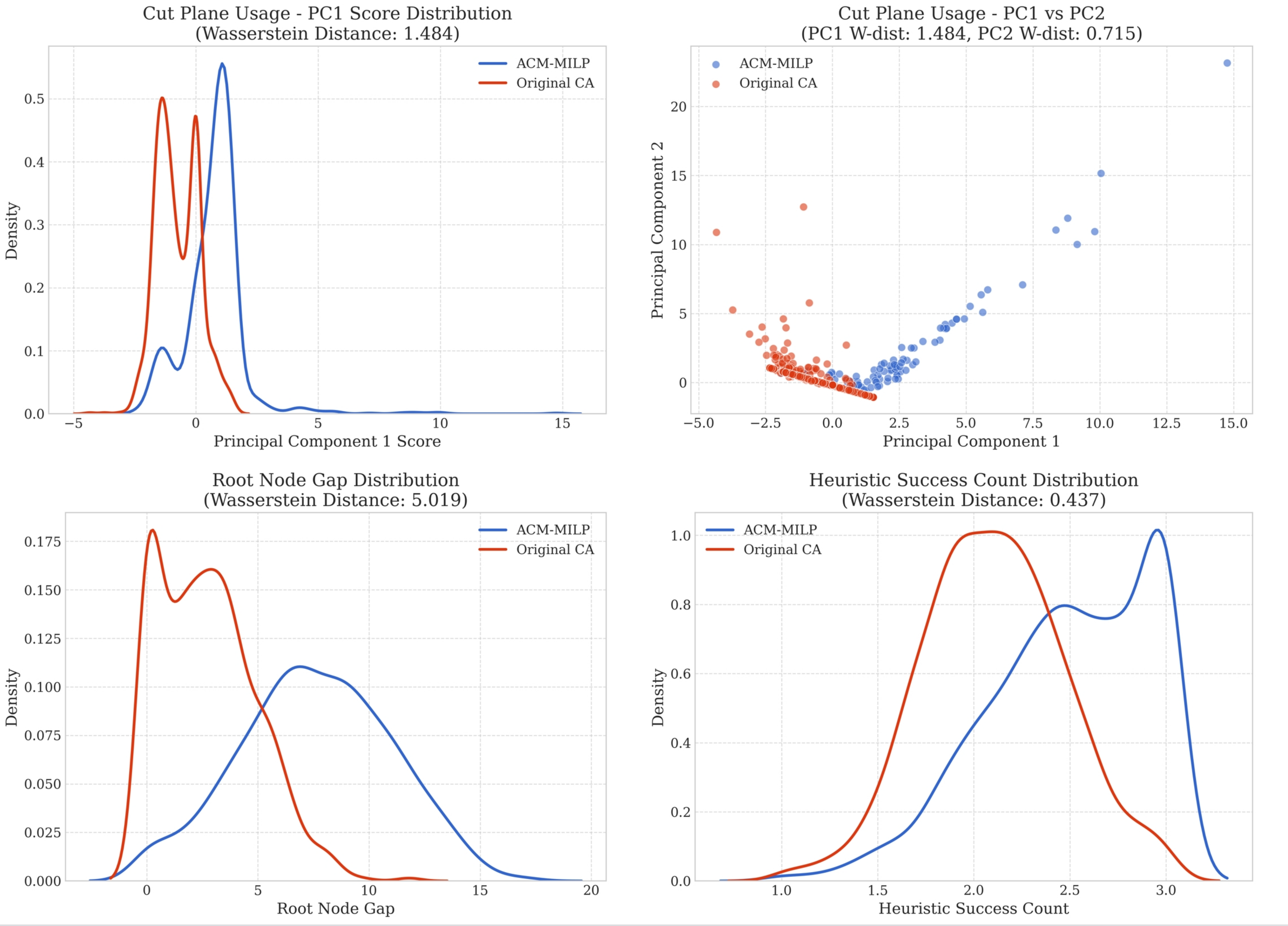}
    \caption{ACM-MILP CA at $\eta = 0.1$ compared with Original Dataset}
    \label{fig:acm_ca_0.1_sif}
\end{figure}

\begin{figure}
    \centering
    \includegraphics[width=1\linewidth]{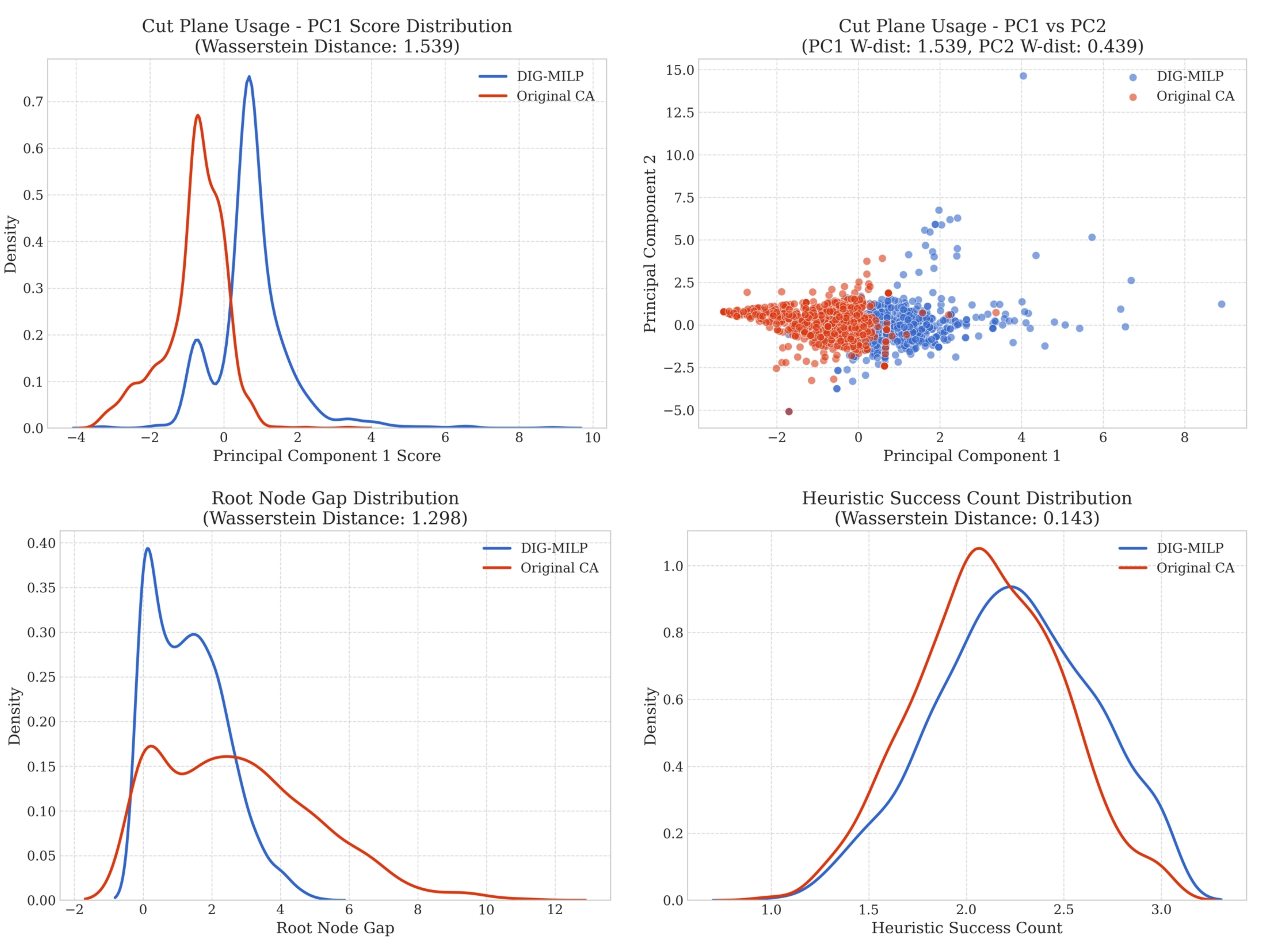}
    \caption{DIG-MILP CA at $\eta = 0.1$ compared with Original Dataset}
    \label{fig:dig_ca_0.1_sif}
\end{figure}

\begin{figure}
    \centering
    \includegraphics[width=1\linewidth]{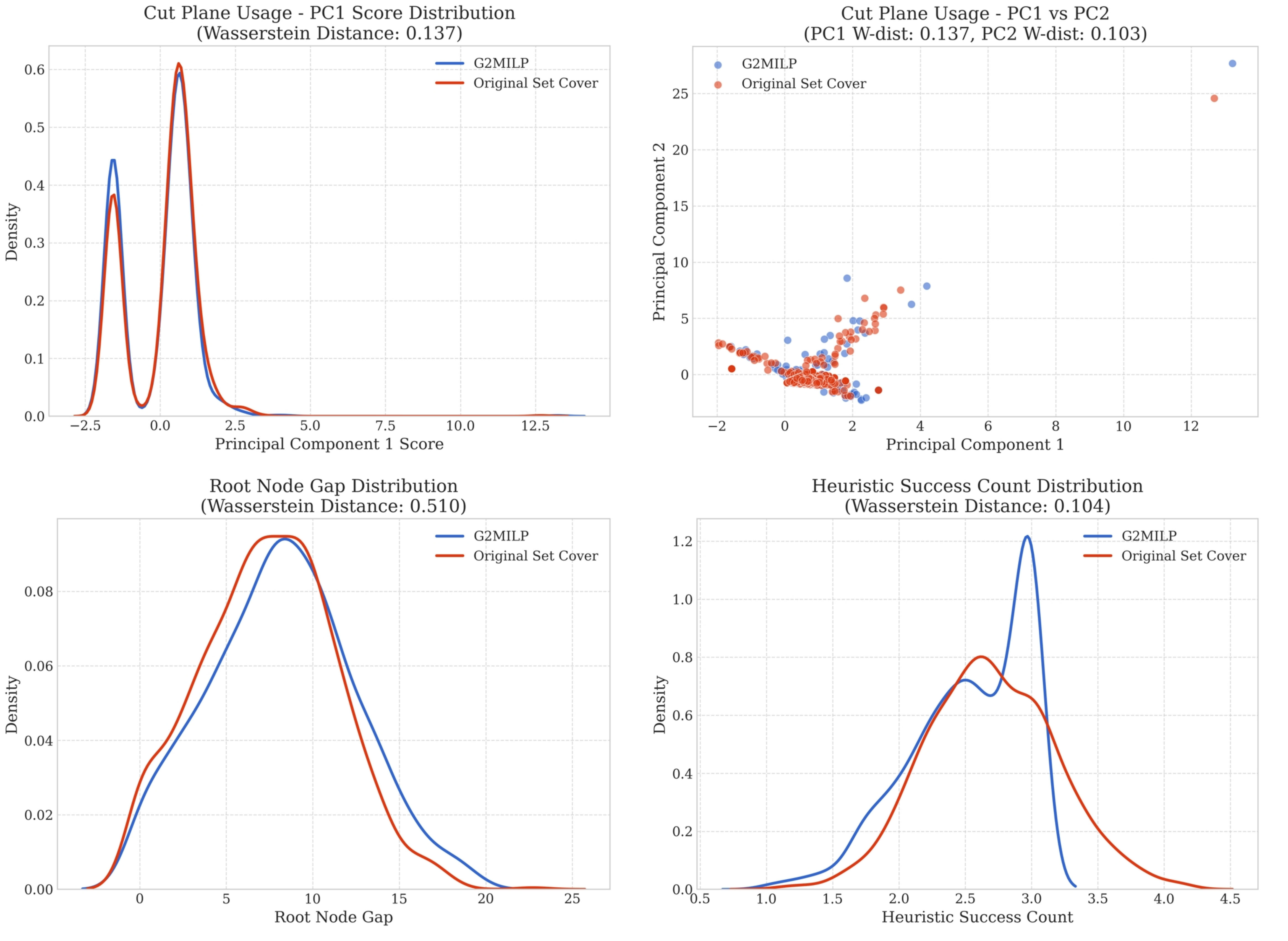}
    \caption{G2MILP SC at $\eta = 0.1$ compared with Original Dataset}
    \label{fig:g2milp_sc_0.1}
\end{figure}

Here are some figures~\ref{fig:acm_mis_0.1_sif} \ref{fig:acm_ca_0.1_sif},\ref{fig:dig_ca_0.1_sif},\ref{fig:g2milp_sc_0.1} showing the results of solver internal features comparison. 

\subsection{Efficiency Test on Super Hard Instances} \label{ml4co_app}

\subsubsection{Experimental Objective}

To validate the practical feasibility and robustness of our solver-internal feature analysis method, we designed an efficiency test.  
The core goal of this experiment is to demonstrate that even under strict computational constraints (i.e., tight time limits) and when facing a well-known set of hard instances, our method can still stably and efficiently extract meaningful feature distributions.  
This supports our key hypothesis that obtaining such deep behavioral features does not require solving instances to optimality, thereby making the evaluation both low-cost and highly efficient.

\subsubsection{Experimental Methodology}

\textbf{Dataset Selection}:  
    We used the \texttt{item\_placement} dataset from the ML4CO competition, which is recognized for its high solving difficulty.  
    A total of 1000 instances were selected for testing.

\textbf{Computational Budget Constraint}:  
    To simulate scenarios with limited computational resources, we imposed a strict time limit of \textbf{120 seconds} for the Gurobi solver on each instance.

\begin{flushleft}
    \textbf{Validation Method}: 
    We employed a split-half validation approach to evaluate the stability of the extracted features:
    \begin{itemize}[leftmargin=*]
        \item The 1000 solver logs were randomly divided into two halves, each containing 500 instances.
        \item We independently extracted solver-internal features from each half, including Root Node Gap, Heuristic Success Count, and Cut Plane Usage.
        \item We then computed and compared the distributional similarity of these features across the two halves. 
        If the random subsets originating from the same underlying distribution exhibit highly similar feature distributions (i.e., low Wasserstein distance), it indicates strong stability and reliability of the feature-extraction method.
    \end{itemize}
\end{flushleft}

\subsubsection{Experimental Results}

The split-half experiment clearly shows that even under strict time constraints, the distributions of solver-internal features extracted from the solving process remain highly stable.  
Key indicators and their 1-Wasserstein distances are summarized in Table~\ref{tab:efficiency-hard}.

\vspace{10pt}

\begin{table}[t!]
\centering
\caption{W-1 distances of key solver-internal features across two random halves}
\label{tab:efficiency-hard}
\begin{tabularx}{\linewidth}{l c X}   
\hline
\textbf{Feature Dimension} & \textbf{W-1 Dist.} & \textbf{Remarks} \\
\hline
Root Node Gap & 0.7613 & Mean difference only 0.04\%; identical standard deviation \\
Heuristic Success Count & 0.7320 & Mean and standard deviation are very close \\
Cut Plane Usage & & Compared on top 3 principal components after PCA \\
\quad -- PC1 & 0.0917 & Extremely low distance indicates highly consistent cut-plane usage pattern \\
\quad -- PC2 & 0.0870 &  \\
\quad -- PC3 & 0.0904 &  \\
\hline
\end{tabularx}
\end{table}

\subsubsection{Conclusion}

This experiment strongly demonstrates the practical value of our framework:
\begin{itemize}[leftmargin=*]
    \item \textbf{High Efficiency and Low Cost}:
    The results confirm that a truncated solving process (120-second timeout) is sufficient to extract stable and representative solver-internal feature distributions.
    This substantially reduces the time and computational cost of evaluating large-scale or extremely hard instance sets.
    \item \textbf{Feature Stability}:
    The very small Wasserstein distances observed for Root Node Gap and Cut Plane Usage indicate that these early-stage solver behavior features are highly stable.
    \item \textbf{Practical Feasibility}:
    The success of this test shows that the GenBench-MILP framework is not only theoretically sound but also practically applicable for evaluating MILP instances that are difficult to solve to optimality within a reasonable time frame and that reflect real-world complexity.
\end{itemize}

\subsection{Tests on Other Solvers} \label{solvers}

\subsubsection{Experimental Objective}

To validate the generality of the GenBench-MILP framework and to investigate potential differences in solver-internal behavioral features, we applied our core solver-internal feature analysis method to two leading open-source solvers: \textbf{SCIP} and \textbf{HiGHS}.  
This experiment is designed to answer two key questions:
\begin{enumerate}
    \item \textit{Can our framework be flexibly adapted to solvers other than Gurobi?}
    \item \textit{Do different solvers exhibit varying stability in their internal behaviors when solving the same instance set?}
\end{enumerate}

\subsubsection{Experimental Methodology}

We adopted the same split-half validation method used in the Gurobi experiments.  
Using the original Independent Set (IS) dataset, we randomly divided the instances into two subsets of 500 instances each.  
SCIP and HiGHS were then used to solve the two subsets separately, extracting three categories of solver-internal features: Root Node Gap, Heuristic Success, and Cut Plane Usage.  
Finally, we computed the 1-Wasserstein distance of these feature distributions between the two subsets to quantify the stability of each solver's internal behavior.  
For a stable solver, the Wasserstein distance between random subsets originating from the same distribution should remain very small.

\subsubsection{Experimental Results and Analysis}

Our results show that, while the framework can be successfully applied to all tested solvers, their internal feature stability varies substantially.

\paragraph{SCIP Solver}

\begin{figure}
    \centering
    \includegraphics[width=1\linewidth]{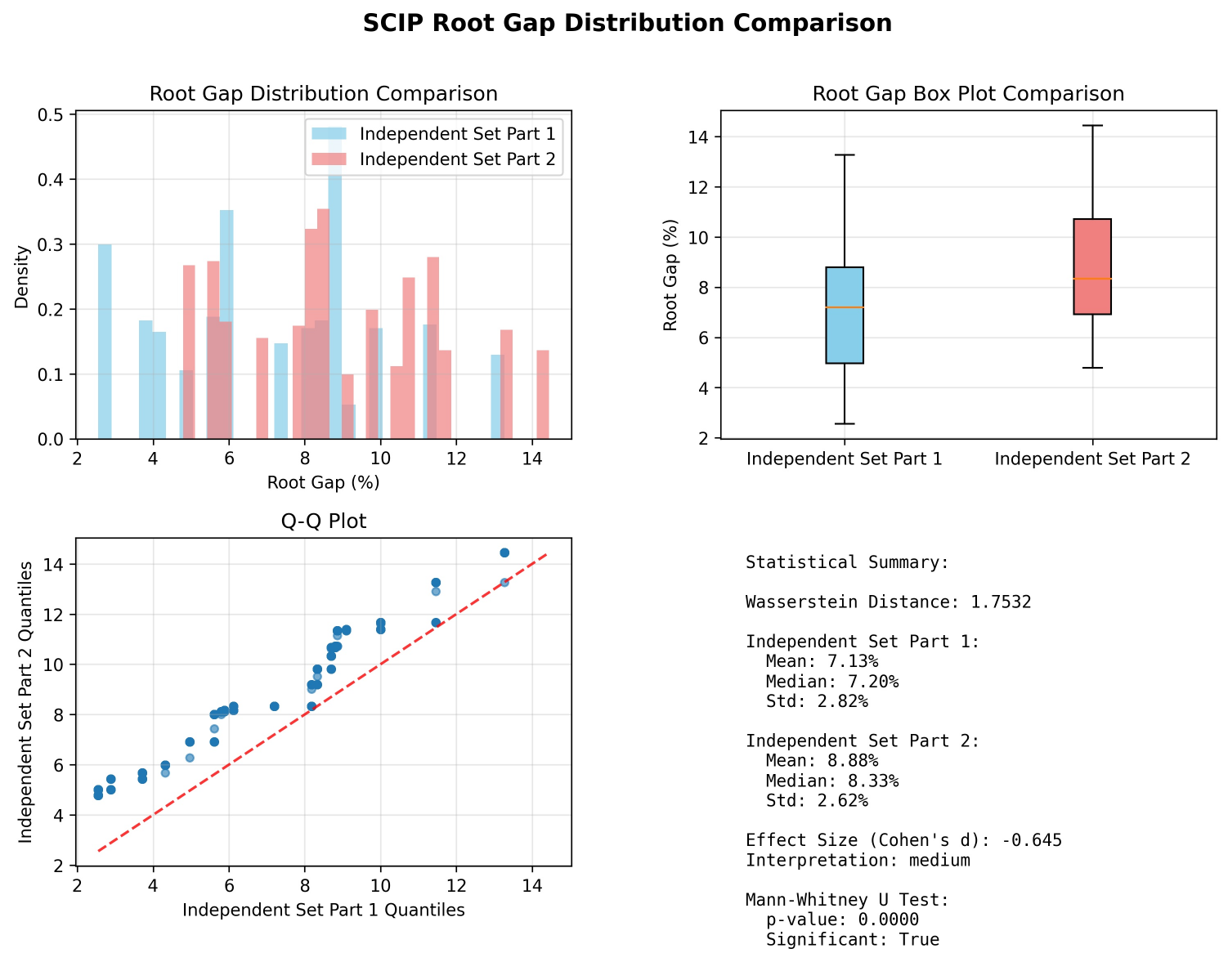}
    \caption{Comparison of Root Node Gap distributions for SCIP in split-half validation. The two distributions represent the two random halves of the instance set.}
    \label{fig:scip_root}
\end{figure}

\begin{figure}
    \centering
    \includegraphics[width=1\linewidth]{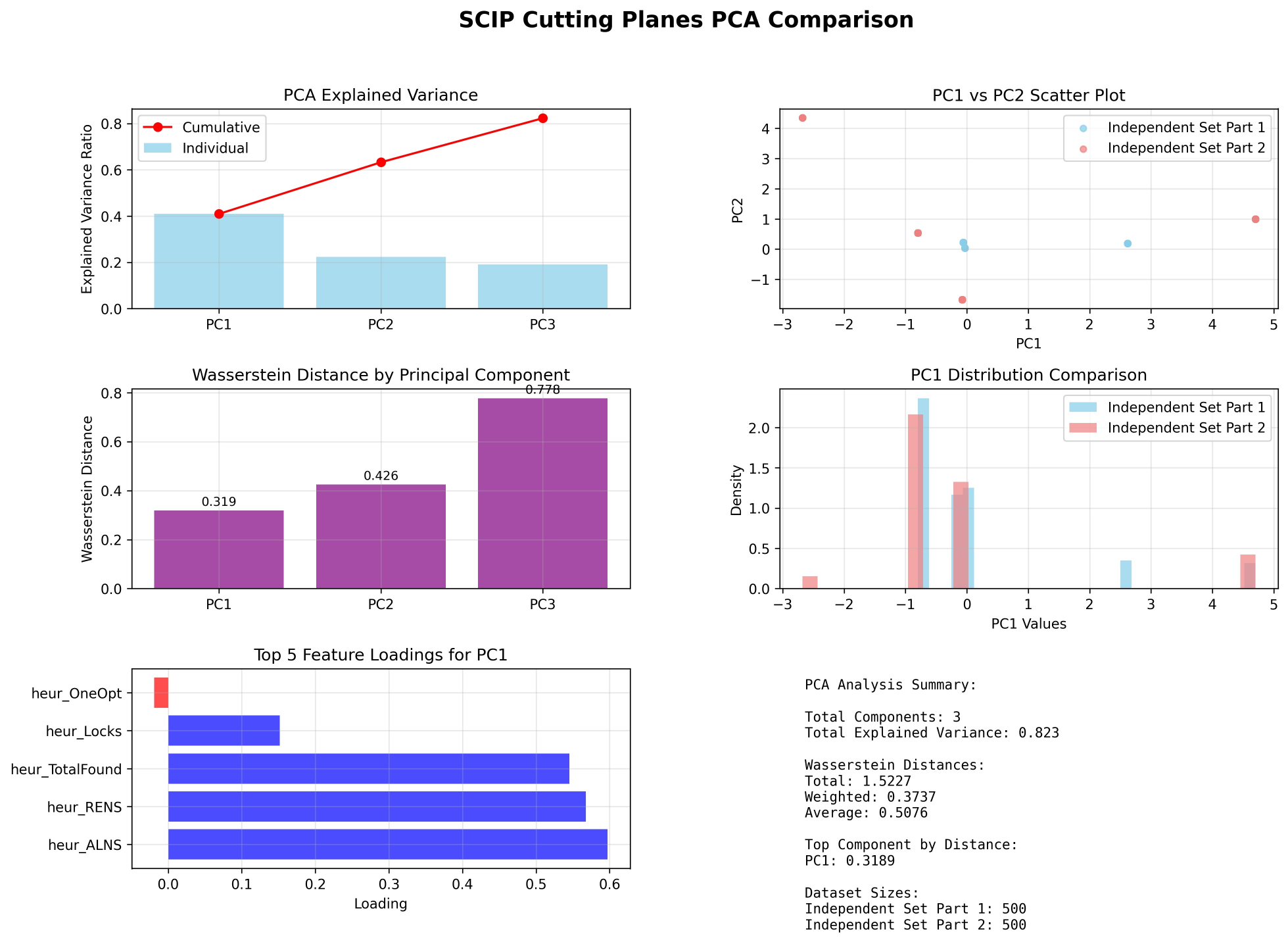}
    \caption{Comparison of Heuristic distributions for SCIP in split-half validation. The two distributions represent the two random halves of the instance set.}
    \label{fig:scip_heur}
\end{figure}

\begin{figure}
    \centering
    \includegraphics[width=1\linewidth]{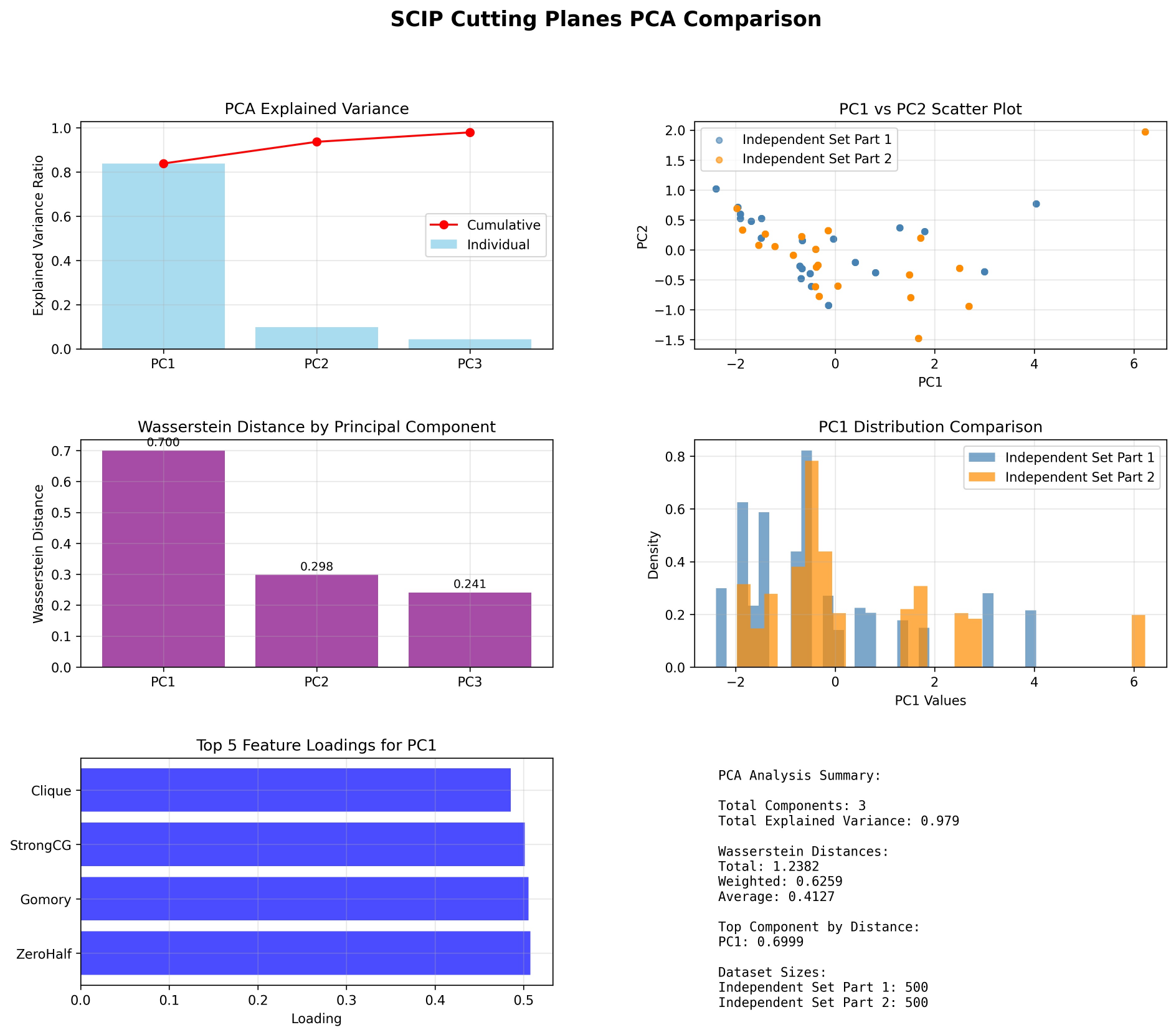}
    \caption{Comparison of cut plane distributions for SCIP in split-half validation. The two distributions represent the two random halves of the instance set.}
    \label{fig:scip_cut}
\end{figure}

Here are some figures~\ref{fig:scip_root}, \ref{fig:scip_heur} and \ref{fig:scip_cut} showing the function of SCIP in spilt-half validation. For SCIP, the split-half experiment yields the following:
\begin{itemize}[leftmargin=*]
    \item \textbf{Root Node Gap}:  
    The two subsets exhibit mean gaps of 7.13\% and 8.88\%, respectively.  
    The Wasserstein distance is 1.753, which is much larger than the 0.1884 observed for Gurobi, indicating lower stability for SCIP in this metric.
    \item \textbf{Heuristic Success}:  
    Principal Component Analysis (PCA) of six heuristic methods shows Wasserstein distances of  
    PC1: 0.319, PC2: 0.426, and PC3: 0.778.
    \item \textbf{Cut Plane Usage}:  
    PCA on four cut-plane types gives Wasserstein distances of  
    PC1: 0.700, PC2: 0.298, and PC3: 0.241.
\end{itemize}

\paragraph{HiGHS Solver}

\begin{figure}
    \centering
    \includegraphics[width=1\linewidth]{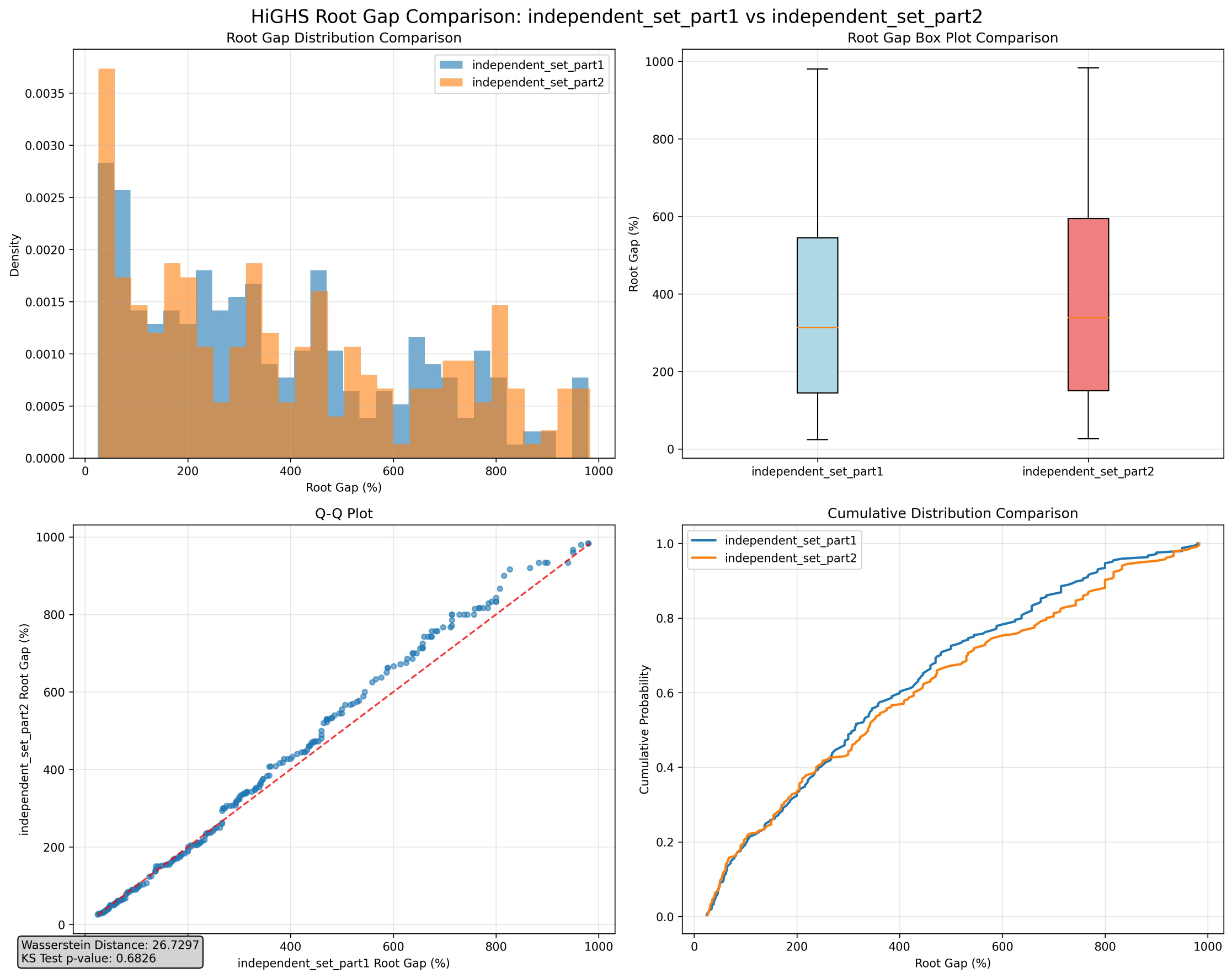}
    \caption{Comparison of root node gap distributions for HiGHS in split-half validation. The two distributions represent the two random halves of the instance set.}
    \label{fig:highs_root}
\end{figure}

\begin{figure}
    \centering
    \includegraphics[width=1\linewidth]{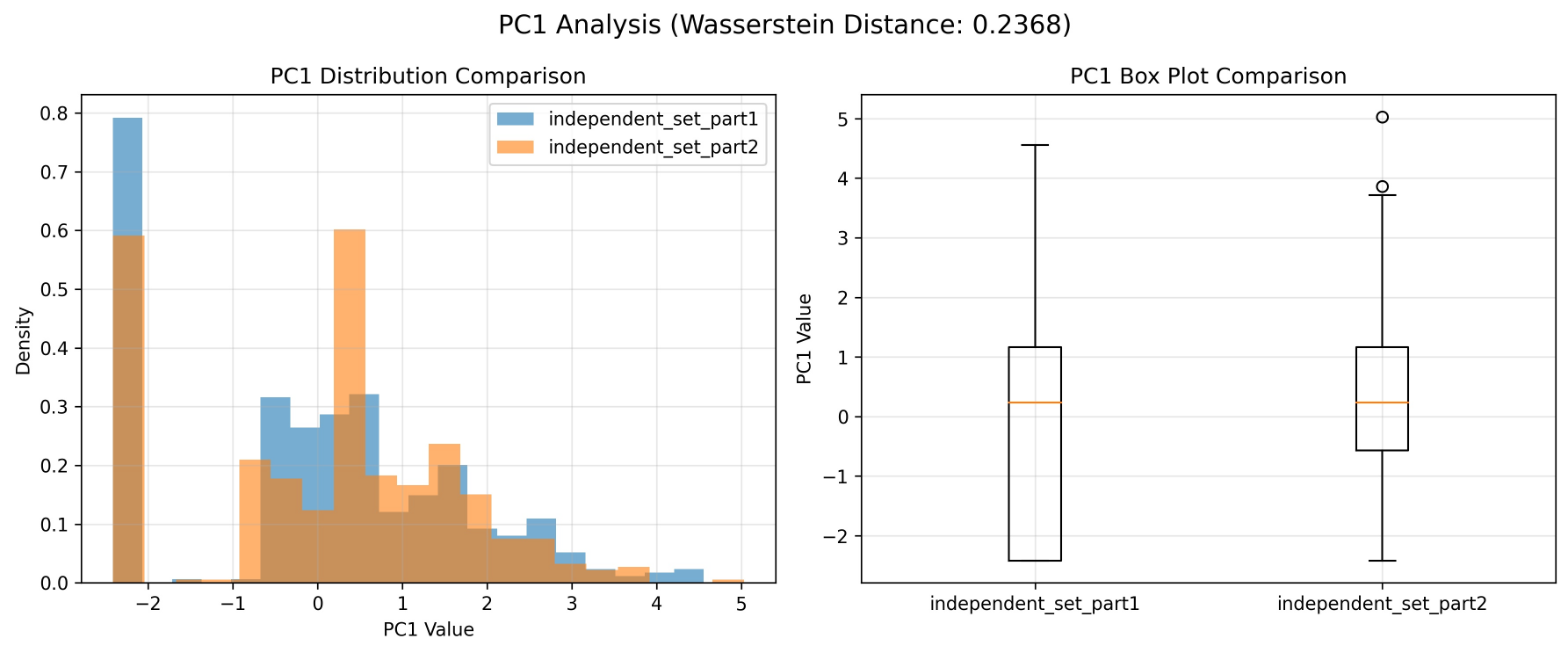}
    \caption{Comparison of heuristic distributions for HiGHS in split-half validation. The two distributions represent the two random halves of the instance set.}
    \label{fig:highs_heur}
\end{figure}

\begin{figure}
    \centering
    \includegraphics[width=1\linewidth]{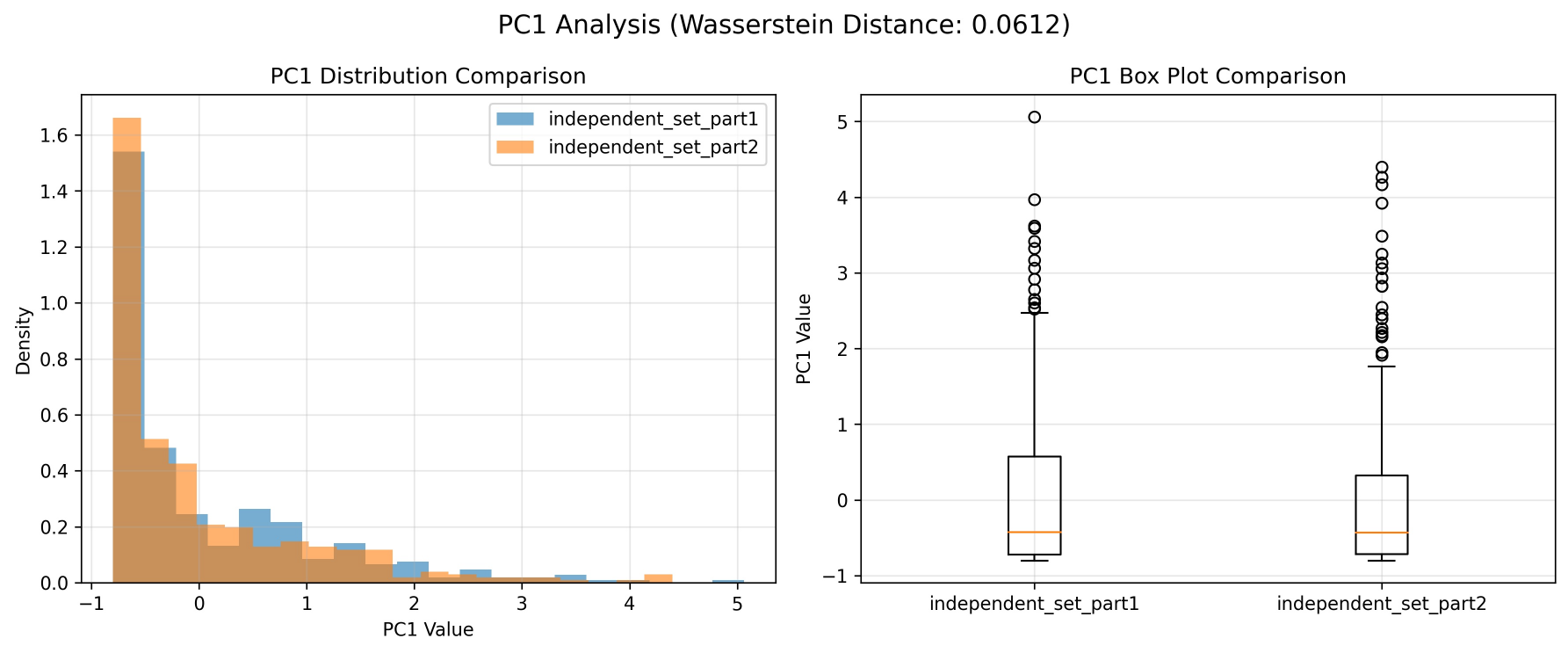}
    \caption{Comparison of cut plane distributions for HiGHS in split-half validation. The two distributions represent the two random halves of the instance set.}
    \label{fig:highs_cut}
\end{figure}

Here are some figures~\ref{fig:highs_root}, \ref{fig:highs_heur} and \ref{fig:highs_cut} showing the function of HiGHS in spilt-half validation. For HiGHS, the differences are even more pronounced:
\begin{itemize} [leftmargin=*]
    \item \textbf{Root Node Gap}:  
    The two subsets have mean gaps of 364.28\% and 386.68\%, respectively.  
    The Wasserstein distance reaches 26.7297, indicating extremely unstable root-node relaxation behavior.
    \item \textbf{Heuristic Success}:  
    PCA analysis of heuristic methods gives Wasserstein distances of  
    PC1: 0.2368 and PC2: 0.1695, with a weighted average of 0.1530, reflecting good stability.
    \item \textbf{Cut Plane Usage}:  
    HiGHS shows very consistent cut-plane usage.  
    PCA reveals that the first principal component explains 100\% of the variance, with a Wasserstein distance of only 0.0612, indicating highly stable cut-plane strategies.
\end{itemize}

\subsubsection{Cross-Solver Comparison}

To facilitate comparison, Table~\ref{tab:cross-solver-wdist2} summarizes key 1-Wasserstein distances from the IS split-half validation for Gurobi, SCIP, and HiGHS.

\vspace{10pt}

\begin{table}[h]
\centering
\caption{1-Wasserstein distances across solvers on IS split-half validation}
\label{tab:cross-solver-wdist2}
\begin{tabular}{lccc}
\hline
\textbf{Feature Dimension} & \textbf{Gurobi (Baseline)} & \textbf{SCIP} & \textbf{HiGHS} \\
\hline
Root Node Gap & \textbf{0.1884} & 1.7530 & \textbf{26.7297} \\
Heuristic (PC1) & \textbf{0.1300} & 0.3190 & 0.2368 \\
Cut Plane (PC1) & 0.2318 & 0.7000 & \textbf{0.0612} \\
\hline
\end{tabular}
\end{table}

\vspace{10pt}

From the table we draw three key conclusions:
\begin{itemize} [leftmargin=*]
    \item \textbf{Framework Generality}:  
    GenBench-MILP can be successfully applied to SCIP and HiGHS, confirming its potential as a general-purpose evaluation tool.
    \item \textbf{Stability Differences Across Solvers}:  
    The most striking observation is the difference in Root Node Gap stability.  
    Gurobi demonstrates exceptionally high stability, SCIP shows moderate instability, and HiGHS exhibits severe instability.  
    This implies that solver strategies (e.g., preprocessing, initial LP relaxations) may have fundamentally different sensitivities to small instance perturbations.
    \item \textbf{Rationale for Choosing Gurobi as Baseline}:  
    These findings strongly justify using Gurobi as the “expert evaluator” in our main study.  
    Accurate and reliable instance-quality assessment demands a solver with highly stable internal behavior.  
    Gurobi’s stability enables precise detection of subtle computational similarities between instances, whereas instability in other solvers can obscure these signals.
\end{itemize}

In summary, although the GenBench-MILP framework is solver-agnostic, the quality of its evaluation results is closely tied to the stability of the solver’s internal strategies.  
State-of-the-art commercial solvers like Gurobi provide a more precise and reliable benchmark for our evaluation framework.  
At the same time, these experiments reveal an additional potential application of GenBench-MILP: as a diagnostic tool to analyze and compare the internal stability and dynamic behavior of different solvers.